\documentclass[journal]{IEEEtran}
\IEEEoverridecommandlockouts         

\usepackage{graphics} 
\usepackage[linesnumbered,ruled]{algorithm2e}
\graphicspath{{img/}}
\usepackage{epsfig} 
\usepackage{amsmath}
\usepackage{changepage}
\usepackage[graphicx]{realboxes}
\usepackage{adjustbox}
\usepackage{footnote}
\usepackage{amssymb}  
\usepackage{pifont}
\usepackage{cite}
\usepackage{tablefootnote} 
\usepackage[makeroom]{cancel}
\usepackage[table]{xcolor} %

\usepackage[center]{subfigure}
\usepackage{multirow}
\usepackage{booktabs}
\usepackage{subfigure}
\usepackage{caption}
\usepackage{epstopdf}
\usepackage{textcomp}
\usepackage{bm}

\usepackage{algpseudocode,algorithm} 
\usepackage{color}
\usepackage{mathptmx}
\usepackage[11pt]{moresize}
\usepackage{amsmath}
\usepackage{hhline}
\usepackage{pdflscape}
\usepackage{balance}
\usepackage{ltablex}
\usepackage{hyperref}
\usepackage{adjustbox}
\usepackage{threeparttablex}
\usepackage{enumitem}
\usepackage{units}
\usepackage{mathtools}

\usepackage[normalem]{ulem}

\usepackage{tikz}
\newcommand\copyrighttext{%
	\scriptsize \textcopyright 2020 IEEE. Personal use of this material is permitted. Permission from IEEE must be obtained for all other uses, in any current or future media, including reprinting/republishing this material for advertising or promotional purposes, creating new collective works, for resale or redistribution to servers or lists, or reuse of any copyrighted component of this work in other works. DOI: 10.1109/TITS.2020.2972974.}
\newcommand\copyrightnotice{%
	\begin{tikzpicture}[remember picture,overlay]
	\node[anchor=south,yshift=10pt] at (current page.south) {\fbox{\parbox{\dimexpr\textwidth-\fboxsep-\fboxrule\relax}{\copyrighttext}}};
	\end{tikzpicture}%
}

\begin{document}

\bstctlcite{IEEE_BSTcontrol}

\title{Deep Multi-modal Object Detection and Semantic Segmentation for Autonomous Driving: Datasets, Methods, and Challenges }

\author{Di Feng$^{\ast,\text{1,2},\dagger}$, Christian Haase-Sch\"utz$^{\ast,\text{3,4}}$, Lars Rosenbaum$^{\text{1}}$, Heinz Hertlein$^{\text{3}}$, Claudius Gl\"aser$^{\text{1}}$, Fabian Timm$^{\text{1}}$, Werner Wiesbeck$^{\text{4}}$, Klaus Dietmayer$^{\text{2}}$
%
\thanks{$^{\ast}$ Di Feng and Christian Haase-Sch\"utz contributed equally to this work.}
\thanks{$^{\text{1}}$ Driver Assistance Systems and Automated Driving, Corporate Research, Robert Bosch GmbH, 71272 Renningen, Germany.} 
\thanks{$^{\text{2}}$ Institute of Measurement, Control and Microtechnology, Ulm University, 89081 Ulm, Germany. 
}
\thanks{$^{\text{3}}$ Engineering Cognitive Systems, Automated Driving, Chassis Systems Control, Robert Bosch GmbH, 74232 Abstatt, Germany.} 
\thanks{$^{\text{4}}$ Institute of Radio Frequency Engineering and Electronics, Karlsruhe Institute of Technology, 76131 Karlsruhe, Germany. 
}
\thanks{$^{\dagger}$ Corresponding author: Di.Feng@de.bosch.com}
}

\maketitle
\begin{abstract}
Recent advancements in perception for autonomous driving are driven by deep learning. In order to achieve robust and accurate scene understanding, autonomous vehicles are usually equipped with different sensors (e.g. cameras, LiDARs, Radars), and multiple sensing modalities can be fused to exploit their complementary properties. In this context, many methods have been proposed for deep multi-modal perception problems. However, there is no general guideline for network architecture design, and questions of ``what to fuse'', ``when to fuse'', and ``how to fuse'' remain open. This review paper attempts to systematically summarize methodologies and discuss challenges for deep multi-modal object detection and semantic segmentation in autonomous driving. To this end, we first provide an overview of on-board sensors on test vehicles, open datasets, and background information for object detection and semantic segmentation in autonomous driving research. We then summarize the fusion methodologies and discuss challenges and open questions. In the appendix, we provide tables that summarize topics and methods. We also provide an interactive online platform to navigate each reference: \url{https://boschresearch.github.io/multimodalperception/}.
\end{abstract}

\IEEEpeerreviewmaketitle

\copyrightnotice

\section{\textbf{Introduction}}\label{sec:intro}
\begin{figure}[tpb]
\centering
	\includegraphics[width=0.96\linewidth]{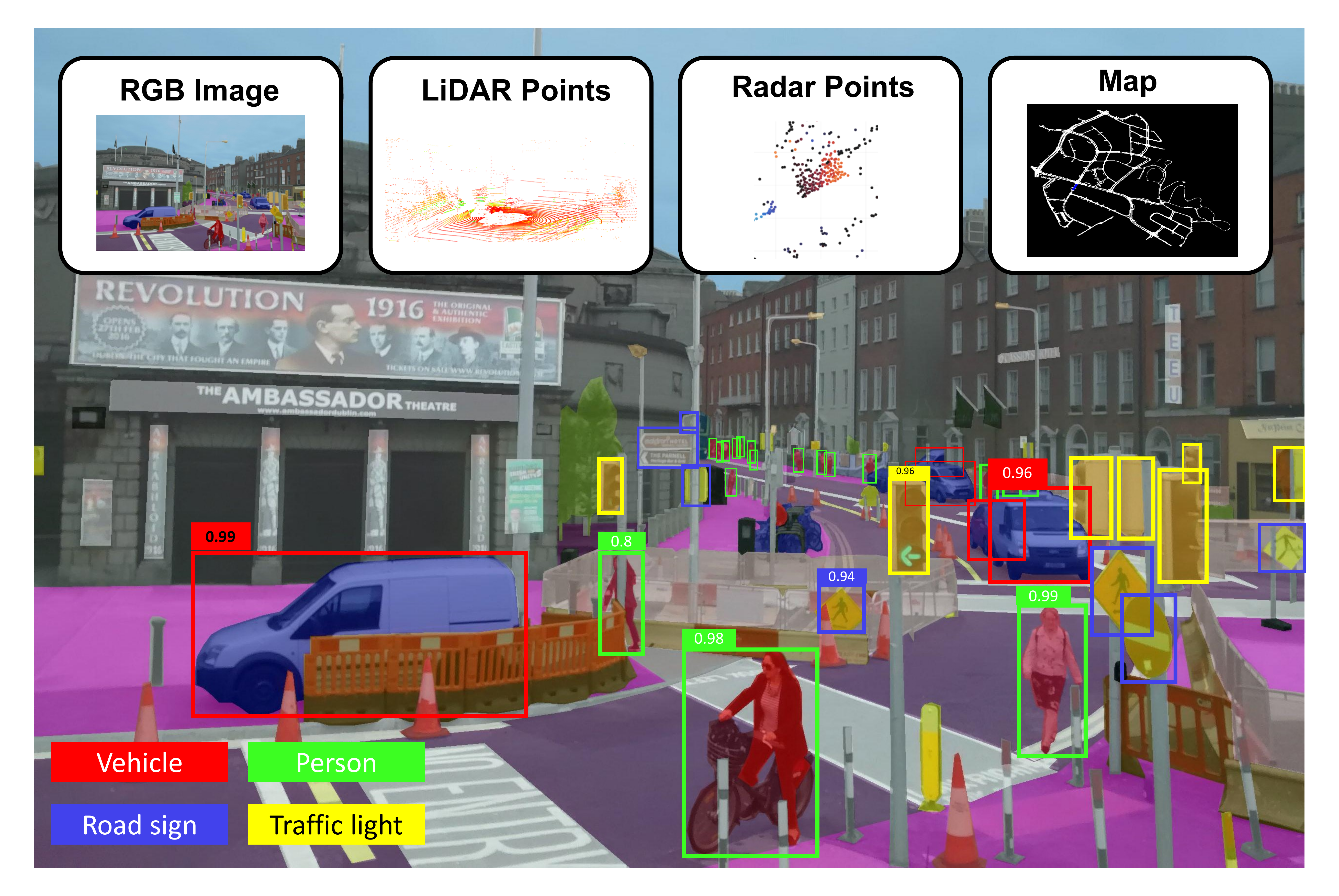} 
	\caption{A complex urban scenario for autonomous driving. The driverless car uses multi-modal signals for perception, such as RGB camera images, LiDAR points, Radar points, and map information. It needs to perceive all relevant traffic participants and objects accurately, robustly, and in real-time. For clarity, only the bounding boxes and classification scores for some objects are drawn in the image. The RGB image is adapted from \cite{neuhold2017mapillary}.}\label{fig:scene}
	\vspace{-0.4cm}
\end{figure}

Significant progress has been made in autonomous driving since the first successful demonstration in the 1980s~\cite{dickmanns1992recursive} and the DARPA Urban Challenge in 2007~\cite{urmson2008autonomous}. It offers high potential to decrease traffic congestion, improve road safety, and reduce carbon emissions~\cite{berger2014autonomous}. However, developing reliable autonomous driving is still a very challenging task. This is because driverless cars are intelligent agents that need to perceive, predict, decide, plan, and execute their decisions in the real world, often in uncontrolled or complex environments, such as the urban areas shown in Fig.~\ref{fig:scene}. A small error in the system can cause fatal accidents. 

\begin{figure}[tbp]
	\centering
		\centering
	\includegraphics[width=1\linewidth]{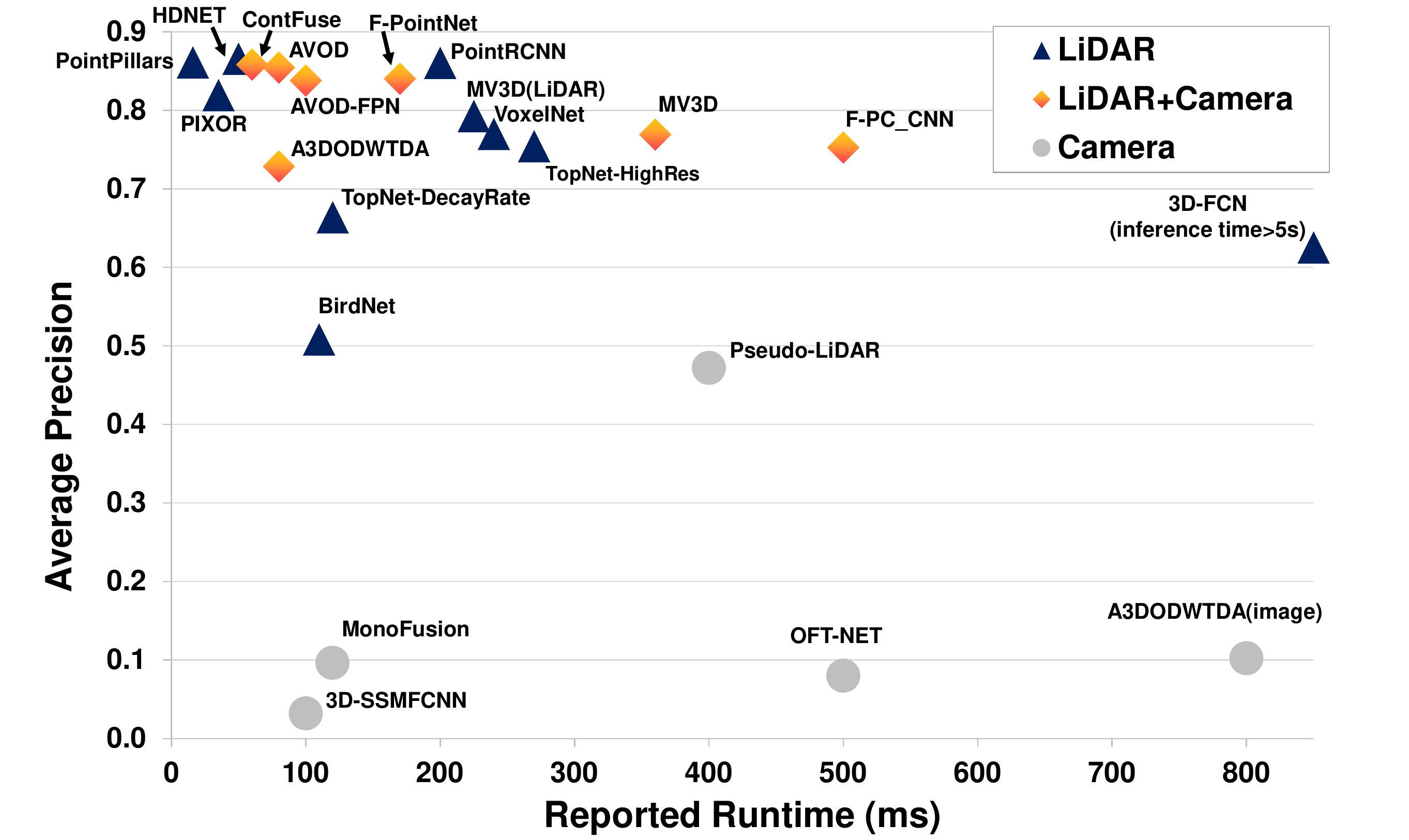}
	\caption{Average precision (AP) vs. runtime. Visualized are deep learning approaches that use LiDAR, camera, or both as inputs for car detection on the KITTI bird's eye view test dataset. Moderate APs are summarized. The results are mainly based on the KITTI leader-board \cite{Geiger2012CVPR} (visited on Apr. 20, 2019). On the leader-board only the published methods are considered.}\label{fig:bev_car_leaderboard}
    \vspace{-2.5mm}
\end{figure}

Perception systems in driverless cars need to be (1). \textit{accurate}: they need to give precise information of driving environments; (2). \textit{robust}: they should work properly in adverse weather, in situations that are not covered during training (open-set conditions), and when some sensors are degraded or even defective; and (3). \textit{real-time}: especially when the cars are driving at high speed. Towards these goals, autonomous cars are usually equipped with multi-modal sensors (e.g. cameras, LiDARs, Radars), and different sensing modalities are fused so that their complementary properties are exploited (cf. Sec. \ref{subsec:sensor}). Furthermore, deep learning has been very successful in computer vision. A deep neural network is a powerful tool for learning hierarchical feature representations given a large amount of data~\cite{lecun2015deep}. In this regard, many methods have been proposed that employ deep learning to fuse multi-modal sensors for scene understanding in autonomous driving. Fig.~\ref{fig:bev_car_leaderboard} shows some recently published methods and their performance on the KITTI dataset~\cite{Geiger2012CVPR}. All methods with the highest performance are based on deep learning, and many methods that fuse camera and LiDAR information produce better performance than those using either LiDAR or camera alone. In this paper, we focus on two fundamental perception problems, namely, \textbf{object detection} and \textbf{semantic segmentation}. In the rest of this paper, we will call them \textbf{deep multi-modal perception} unless mentioned otherwise. 

When developing methods for deep multi-modal object detection or semantic segmentation, it is important to consider the input data: Are there any multi-modal datasets available and how is the data labeled (cf. Tab.~\ref{tab:datasets})? Do the datasets cover diverse driving scenarios (cf. Sec.~\ref{subsubsec:data_diversity})? Is the data of high quality (cf. Sec.~\ref{subsubsec:data_quality})? Additionally, we need to answer several important questions on designing the neural network architecture: Which modalities should be combined via fusion, and how to represent and process them properly (``What to fuse" cf. Sec.~\ref{subsubsec:challenge:what_to_fuse})? Which fusion operations and methods can be used (``How to fuse" cf. Sec.~\ref{subsubsec:challenge:how_to_fuse})? Which stage of feature representation is optimal for fusion (``When to fuse" cf. Sec.~\ref{subsubsec:challenge:how_to_fuse})? 

\subsection{Related Works}
Despite the fact that many methods have been proposed for deep multi-modal perception in autonomous driving, there is no published summary examining available multi-modal datasets, and there is no guideline for network architecture design. Yin \textit{et al.}~\cite{yin2017use} summarize $27$ datasets for autonomous driving that were published between 2006 and 2016, including the datasets recorded with a single camera alone or multiple sensors. However, many new multi-modal datasets have been released since 2016, and it is worth summarizing them. Ramachandram \textit{et al.}~\cite{ramachandram2017deep} provide an overview on deep multi-modal learning, and mention its applications in diverse research fields, such as robotic grasping and human action recognition. Janai \textit{et al.}~\cite{janai2017computer} conduct a comprehensive summary on computer vision problems for autonomous driving, such as scene flow and scene construction. Recently, Arnold \textit{et al.}~\cite{arnold2019survey} survey the 3D object detection problem in autonomous driving. They summarize methods based on monocular images or point clouds, and briefly mention some works that fuse vision camera and LiDAR information. 

\subsection{Contributions}
To the best of our knowledge, there is no survey that focuses on deep multi-modal object detection (2D or 3D) and semantic segmentation for autonomous driving, which makes it difficult for beginners to enter this research field. Our review paper attempts to narrow this gap by conducting a summary of newly-published datasets (2013-2019), and fusion methodologies for deep multi-modal perception in autonomous driving, as well as by discussing the remaining challenges and open questions. 

We first provide background information on multi-modal sensors, test vehicles, and modern deep learning approaches in object detection and semantic segmentation in Sec.~\ref{sec:setup}. We then summarize multi-modal datasets and perception problems in Sec.~\ref{sec:dataset} and Sec.~\ref{sec:perception}, respectively. Sec.~\ref{sec:methodology} summarizes the fusion methodologies regarding ``what to fuse'', ``when to fuse'' and ``how to fuse''. Sec.~\ref{sec:challenge} discusses challenges and open questions when developing deep multi-modal perception systems in order to fulfill the requirements of ``accuracy'', ``robustness'' and ``real-time'', with a focus on data preparation and fusion methodology. We highlight the importance of data diversity, temporal and spatial alignment, and labeling efficiency for multi-modal data preparation. We also highlight the lack of research on fusing Radar signals, as well as the importance of developing fusion methodologies that tackle open dataset problems or increase network robustness. Sec.~\ref{sec:discussion} concludes this work. In addition, we provide an interactive online platform for navigating topics and methods for each reference. The platform can be found here: \url{https://boschresearch.github.io/multimodalperception/}.

\section{\textbf{Background}}\label{sec:setup}
This section provides the background information for deep multi-modal perception in autonomous driving. First, we briefly summarize typical automotive sensors, their sensing modalities, and some vehicles for test and research purposes. Next, we introduce deep object detection and semantic segmentation. Since deep learning has most-commonly been applied to image-based signals, here we mainly discuss image-based methods. We will introduce other methods that process LiDAR and Radar data in Sec.~\ref{subsec:what_to_fuse}. For a more comprehensive overview on object detection and semantic segmentation, we refer the interested reader to the review papers~\cite{liu2018deep,garcia2017review}. For a complete review of computer vision problems in autonomous driving (e.g. optical flow, scene reconstruction, motion estimation), cf.~\cite{janai2017computer}.

\subsection{Sensing Modalities for Autonomous Driving} \label{subsec:sensor}
\subsubsection{Visual and Thermal Cameras} 
Images captured by visual and thermal cameras can provide detailed texture information of a vehicle's surroundings. While visual cameras are sensitive to lighting and weather conditions, thermal cameras are more robust to daytime/nighttime changes as they detect infrared radiation that relates to heat from objects. However, both types of cameras however cannot directly provide depth information. 

\subsubsection{LiDARs}
LiDARs (Light Detection And Ranging) give accurate depth information of the surroundings in the form of 3D points. They measure reflections of laser beams which they emit with a certain frequency. LiDARs are robust to different lighting conditions, and less affected by various weather conditions such as fog and rain than visual cameras. However, typical LiDARs are inferior to cameras for object classification since they cannot capture the fine textures of objects, and their points become sparse with distant objects. Recently, flash LiDARs were developed which can produce detailed object information similar to camera images. Frequency Modulated Continuous Wave (FMCW) LiDARs can provide velocity information.

\subsubsection{Radars} 
Radars (Radio Detection And Ranging) emit radio waves to be reflected by an obstacle, measures the signal runtime, and estimates the object's radial velocity by the Doppler effect. They are robust against various lighting and weather conditions, but classifying objects via Radars is very challenging due to their low resolution. Radars are often applied in adaptive cruise control (ACC) and traffic jam assistance systems~\cite{bengler2014three}.

\subsubsection{Ultrasonics} 
Ultrasonic sensors send out high-frequency sound waves to measure the distance to objects. They are typically applied for near-range object detection and in low speed scenarios, such as automated parking~\cite{bengler2014three}. Due to the sensing properties, Ultrasonics are largely affected by air humidity, temperature, or dirt.  

\subsubsection{GNSS and HD Maps} 
GNSS (Global Navigation Satellite Systems) provide accurate 3D object positions by a global satellite system and the receiver. Examples of GNSS are GPS, Galileo and GLONASS. First introduced to automotive as navigation tools in driver assistance functions \cite{bengler2014three}, currently GNSS is also used together with HD Maps for path planning and ego-vehicle localization for autonomous vehicles.

\subsubsection{IMU and Odometers}
Unlike sensors discussed above which capture information in the external environment (i.e. ``exteroceptive sensors"), Inertial Measurement Units (IMU) and odometers provide vehicles' internal information (i.e. ``proprioceptive sensors")~\cite{bengler2014three}. IMU measure the vehicles' accelerations and rotational rates, and odometers the odometry. They have been used in vehicle dynamic driving control systems since the 1980s. Together with the exteroceptive sensors, they are currently used for accurate localization in autonomous driving.

\subsection{Test Vehicle Setup}\label{subsec:testvehicle}
\begin{figure}[tpb]
    \centering
    \begin{minipage}{0.96\linewidth}
	\centering
	\subfigure[]{\label{fig:cmu}\includegraphics[width=0.48\textwidth]{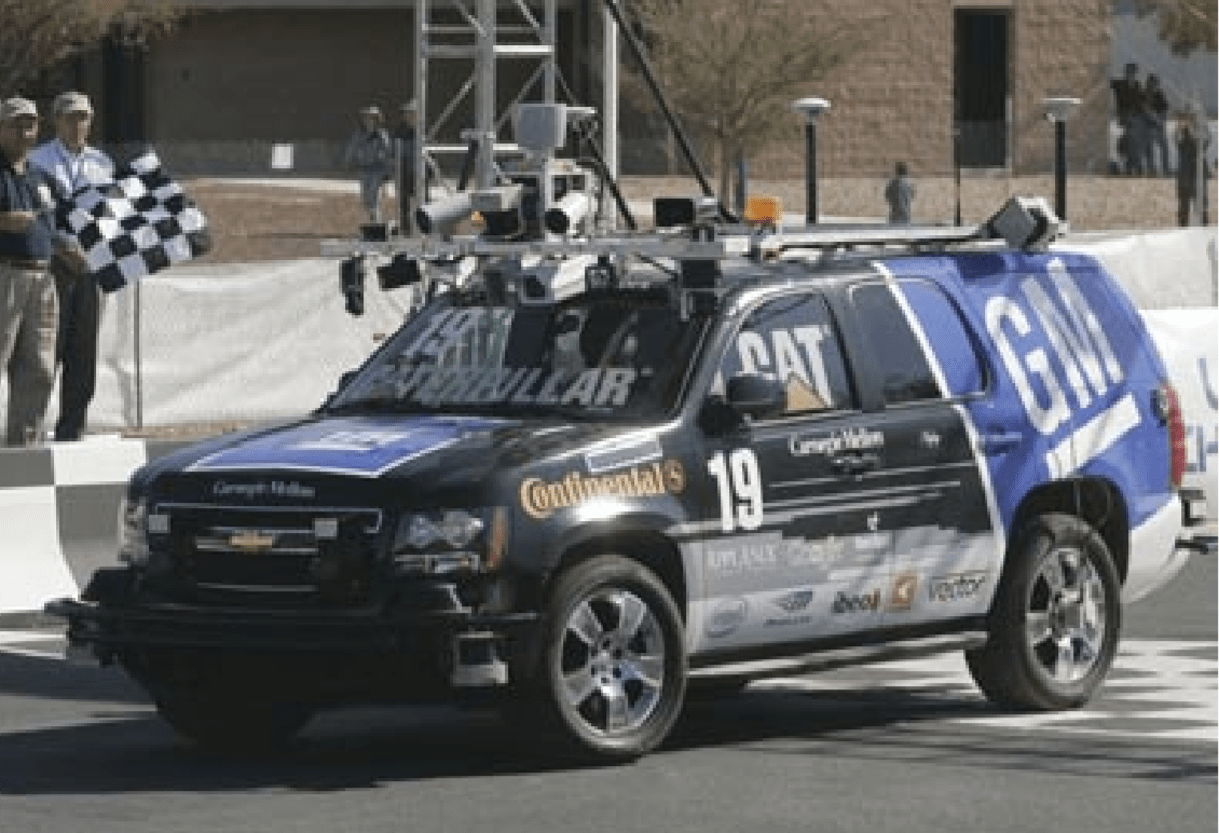}}
	\subfigure[]{\label{fig:google}\includegraphics[width=0.48\textwidth]{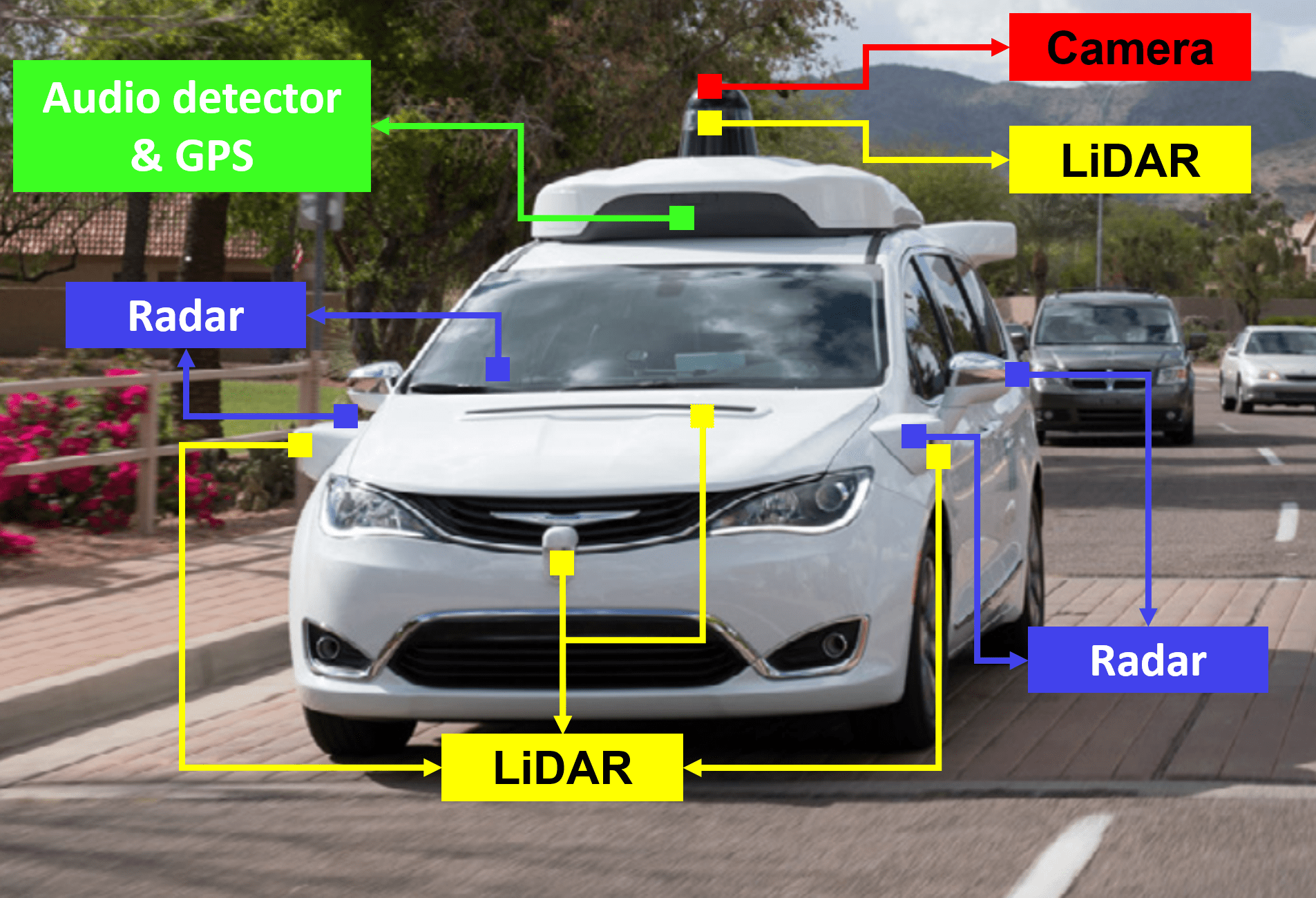}}
    \end{minipage}
    \caption{(a) The Boss autonomous car at DARPA 2007~\cite{urmson2008autonomous}, (b) Waymo self-driving car~\cite{waymo2017autonomous}.}
\end{figure}

Equipped with multiple sensors introduced in Sec.~\ref{subsec:sensor}, many autonomous driving tests have been conducted. For example, the Tartan Racing Team developed an autonomous vehicle called ``Boss" and won the DARPA Urban Challenge in 2007 (cf. Fig.~\ref{fig:cmu})~\cite{urmson2008autonomous}. The vehicle was equipped with a camera and several Radars and LiDARs. Google (Waymo) has tested their driverless cars in more than $20$ US cities driving $8$ million miles on public roads (cf. Fig.~\ref{fig:google})~\cite{waymo2017autonomous}; BMW has tested autonomous driving on highways around Munich since 2011~\cite{aeberhard2015experience}; Daimler mounted a stereo camera, two mono cameras, and several Radars on a Mercedes Benz S-Class car to drive autonomously on the Bertha Benz memorial route in 2013~\cite{ziegler2014making}. Our interactive online platform provides a detailed description for more autonomous driving tests, including Uber, Nvidia, GM Cruise, Baidu Apollo, as well as their sensor setup.

Besides driving demonstrations, real-world datasets are crucial for autonomous driving research. In this regard, several research projects use \textit{data} vehicles with multi-modal sensors to build open datasets. These data vehicles are usually equipped with cameras, LiDARs and GPS/IMUs to collect images, 3D point clouds, and vehicle localization information. Sec.~\ref{sec:dataset} provides an overview of multi-modal datasets in autonomous driving. 

\subsection{Deep Object Detection}\label{subsec:deep_object_detection}
Object detection is the task of recognizing and localizing multiple objects in a scene. Objects are usually recognized by estimating a classification probability and localized with bounding boxes (cf. Fig.~\ref{fig:scene}). Deep learning approaches have set the benchmark on many popular object detection datasets, such as PASCAL VOC~\cite{pascal-voc-2007} and COCO~\cite{lin2014microsoft}, and have been widely applied in autonomous driving, including detecting traffic lights~\cite{weber2016deeptlr,muller2018detecting,bach2017multi,behrendt2017deep}, road signs~\cite{zhu2016traffic,lee2018simultaneous,luo2018traffic}, people~\cite{zhang2018towards,zhang2016faster,chen20183d}, or vehicles~\cite{li20163d,li2016,chen2016monocular,fang2017fine,mousavian20173d}, to name a few. State-of-the-art deep object detection networks follow one of two approaches: the two-stage or the one-stage object detection pipelines. Here we focus on image-based detection.

\subsubsection{Two-stage Object Detection}
\begin{figure}[tbp]
	\centering
		\centering
	\includegraphics[width=0.96\linewidth]{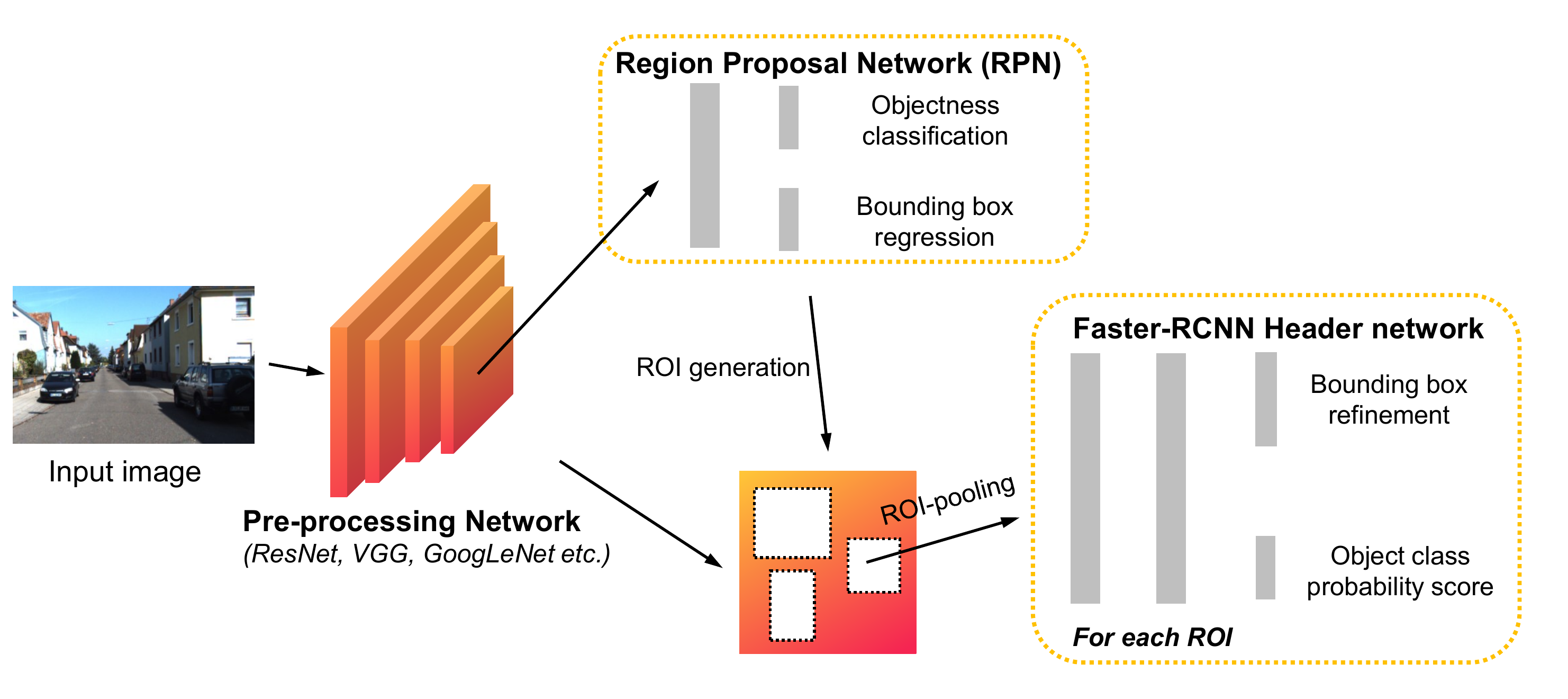}
	\caption{The Faster R-CNN object detection network. It consists of three parts: a pre-processing network to extract high-level image features, a Region Proposal Network (RPN) that produces region proposals, and a Faster-RCNN head which fine-tunes each region proposal.}\label{fig:object_detection}
\end{figure}

In the first stage, several class-agnostic object candidates called regions of interest (ROI) or region proposals (RP) are extracted from a scene. Then, these candidates are verified, classified, and refined in terms of classification scores and locations. OverFeat~\cite{sermanet2013overfeat} and R-CNN~\cite{girshick2014rich} are among pioneering works that employ deep learning for object detection. In these works, ROIs are first generated by the sliding window approach (OverFeat~\cite{sermanet2013overfeat}) or selective search (R-CNN~\cite{girshick2014rich}) and then advanced into a regional CNN to extract features for object classification and bounding box regression. SPPnet~\cite{he2015spatial} and Fast-RCNN~\cite{girshick2015fast} propose to obtain regional features directly from global feature maps by applying a larger CNN (e.g. VGG~\cite{simonyan2014very}, ResNet~\cite{he2016deep}, GoogLeNet~\cite{szegedy2015going}) on the whole image. Faster R-CNN~\cite{ren2015faster} unifies the object detection pipeline and adopts the Region Proposal Network (RPN), a small fully-connected network, to slide over the high-level CNN feature maps for ROI generation (cf. Fig.~\ref{fig:object_detection}). Following this line, R-FCN~\cite{dai2016r} proposes to replace fully-connected layers in an RPN with convolutional layers and builds a fully-convolutional object detector. 

\subsubsection{One-stage Object Detection}
This method aims to map the feature maps directly to bounding boxes and classification scores via a single-stage, unified CNN model. For example, MultiBox~\cite{szegedy2013deep} predicts a binary mask from the entire input image via a CNN and infers bounding boxes at a later stage. YOLO~\cite{redmon2016you} is a more complete unified detector which regresses the bounding boxes directly from the CNN model. SSD~\cite{liu2016ssd} handles objects with various sizes by regressing multiple feature maps of different resolution with small convolutional filters to predict multi-scale bounding boxes. 

In general, two-stage object detectors like Faster-RCNN tend to achieve better detection accuracy due to the region proposal generation and refinement paradigm. This comes with the cost of higher inference time and more complex training.  Conversely, one-stage object detectors are faster and easier to be optimized, yet under-perform compared to two-stage object detectors in terms of accuracy. Huang \textit{et al.}~\cite{huang2017speed} systematically evaluate the speed/accuracy trade-offs for several object detectors and backbone networks. 
\subsection{Deep Semantic Segmentation}\label{subsec:deep_segmentation}
The target of semantic segmentation is to partition a scene into several meaningful parts, usually by labeling each pixel in the image with semantics (pixel-level semantic segmentation) or by simultaneously detecting objects and doing per-instance per-pixel labeling (instance-level semantic segmentation). Recently, panoptic segmentation~\cite{kirillov2018panoptic} is proposed to unify pixel-level and instance-level semantic segmentation, and it starts to get more attentions for autonomous driving~\cite{kirillov2019panoptic, xiong2019upsnet, porzi2019seamless}. Though semantic segmentation was first introduced to process camera images, many methods have been proposed for segmenting LiDAR points as well (e.g.~\cite{wu2017squeezeseg,caltagirone2017fast,huang2018recurrent,dewan2017deep,dewan2019deeptemporalseg,milioto2019iros}).

Many datasets have been published for semantic segmentation, such as Cityscape~\cite{cordts2016cityscapes}, KITTI~\cite{Geiger2012CVPR}, Toronto City~\cite{wang2017torontocity}, Mapillary Vistas~\cite{MVD2017}, and ApolloScape~\cite{apolloscape_arXiv_2018}. These datasets advance the deep learning research for semantic segmentation in autonomous driving. For example,~\cite{schneider2017multimodal,badrinarayanan2017segnet,dewan2017deep} focus on pixel-wise semantic segmentation for multiple classes including road, car, bicycle, column-pole, tree, sky, etc;~\cite{caltagirone2017fast} and \cite{teichmann2016multinet} concentrate on road segmentation; and~\cite{he2017mask,uhrig2018box2pix,wu2017squeezeseg} deal with instance segmentation for various traffic participants.  

Similar to object detection introduced in Sec.~\ref{subsec:deep_object_detection}, semantic segmentation can also be classified into two-stage and one-stage pipelines. In the two-stage pipeline, region proposals are first generated and then fine-tuned mainly for instance-level segmentation (e.g. R-CNN~\cite{gupta2014learning}, SDS~\cite{hariharan2014simultaneous}, Mask-RCNN~\cite{he2017mask}). A more common way for a semantic segmentation is the one-stage pipeline based on a Fully Convolutional Network (FCN) originally proposed by Long \textit{et al.}~\cite{long2015fully}. In this work, the fully-connected layers in a CNN classifier for predicting classification scores are replaced with convolutional layers to produce coarse output maps. These maps are then up-sampled to dense pixel labels by backwards convolution (i.e. deconvolution). Kendall \textit{et al.}~\cite{badrinarayanan2017segnet} extend FCN by introducing an encoder-decoder CNN architecture. The encoder serves to produce hierarchical image representations with a CNN backbone such as VGG or ResNet (removing fully-connected layers). The decoder, conversely, restores these low-dimensional features back to original resolution by a set of upsampling and convolution layers. The restored feature maps are finally used for pixel-label prediction. 

Global image information provides useful context cues for semantic segmentation. However, vanilla CNN structures only focus on local information with limited receptive fields. In this regard, many methods have been proposed to incorporate global information, such as dilated convolutions~\cite{chen2018deeplab,paszke2016enet}, multi-scale prediction~\cite{roy2016multi}, as well as adding Conditional Random Fields (CRFs) as post-processing step~\cite{zheng2015conditional}.

Real-time performance is important in autonomous driving applications. However, most works only focus on segmentation accuracy. In this regard, Siam \textit{et al.}~\cite{siam2018comparative} made a comparative study on the real-time performance among several semantic segmentation architectures, regarding the operations (GFLOPs) and the inference speed (fps).
\section{\textbf{Multi-modal Datasets}}\label{sec:dataset}
Most deep multi-modal perception methods are based on supervised learning. Therefore, multi-modal datasets with labeled ground-truth are required for training such deep neural networks. In the following, we summarize several real-world datasets published since 2013, regarding sensor setups, recording conditions, dataset size and labels (cf. Tab.~\ref{tab:datasets}). Note that there exist some virtual multi-modal datasets generated from game engines. We will discuss them in Sec.~\ref{subsubsec:data_diversity}.

\subsection{Sensing Modalities}
All reviewed datasets include RGB camera images. In addition,~\cite{RobotCarDatasetIJRR,Geiger2013IJRR,Geiger2012CVPR,apolloscape_arXiv_2018,blanco2014malaga,jung2016multi,chen2018lidar, 360LiDARTracking_ICRA_2019, blvdICRA2019, lyft2019, Chang_2019_CVPR, pandaset, waymo_open_dataset, RadarRobotCarDatasetArXiv, astar-3d, aev2019, braun2019eurocity, nuscenes2019} provide LiDAR point clouds, and~\cite{hwang2015multispectral,takumi2017multispectral,ha2017mfnet} thermal images. The KAIST Multispectral Dataset~\cite{choi2018kaist} provides both thermal images and LiDAR data. Bus data is included additionally in \cite{aev2019}. Only the very recently nuScenes~\cite{nuscenes2019}, Oxford Radar RobotCar~\cite{RadarRobotCarDatasetArXiv} and Astyx HiRes2019 Datasets~\cite{meyer2019euma} provide Radar data. 

\subsection{Recording Conditions}
Even though the KITTI dataset~\cite{Geiger2013IJRR} is widely used for autonomous driving research, the diversity of its recording conditions is relatively low: it is recorded in Karlsruhe - a mid-sized city in Germany, only during daytime and on sunny days. Other reviewed datasets such as \cite{aev2019, braun2019eurocity, Chang_2019_CVPR, nuscenes2019, 360LiDARTracking_ICRA_2019, apolloscape_arXiv_2018, chen2018lidar} are recorded in more than one location. To increase the diversity of lighting conditions,~\cite{hwang2015multispectral,takumi2017multispectral,ha2017mfnet,apolloscape_arXiv_2018, blvdICRA2019,waymo_open_dataset, Chang_2019_CVPR, astar-3d, braun2019eurocity, nuscenes2019, Chang_2019_CVPR, lyft2019} collect data in both daytime and nighttime, and \cite{choi2018kaist} considers various lighting conditions throughout the day, including sunrise, morning, afternoon, sunset, night, and dawn. The Oxford Dataset~\cite{RobotCarDatasetIJRR} and the Oxford Radar RobotCar Dataset \cite{RadarRobotCarDatasetArXiv} are collected by driving the car around the Oxford area during the whole year. It contains data under different weather conditions, such as heavy rain, night, direct sunlight and snow. Other datasets containing diverse weather conditions are \cite{astar-3d, braun2019eurocity, nuscenes2019, apolloscape_arXiv_2018}. In \cite{kondermann2014}, LiDAR is used as a reference sensor for generating ground-truth, hence we do not consider it a multi-modal dataset. However the diversity in the recording conditions is large, ranging from dawn to night, as well as reflections, rain and lens flare. The cross-season dataset \cite{larsson2019cross} emphasizes the importance of changes throughout the year. However, it only provides camera images and labels for semantic segmentation. Similarly, the visual localization challenge and the corresponding benchmark \cite{sattler2018benchmarking} cover weather and season diversity (but no new multi-modal dataset is introduced). The recent Eurocity dataset \cite{braun2019eurocity} is the most diverse dataset we have reviewed. It is recorded in different cities from several European countries. All seasons are considered, as well as weather and daytime diversity. To date, the dataset is camera-only and other modalities (e.g. LiDARs) are announced.

\subsection{Dataset Size}
The dataset size ranges from only 1,569 frames up to over 11 million frames. The largest dataset with ground-truth labels that we have reviewed is the nuScenes Dataset~\cite{nuscenes2019} with nearly 1,4M frames. Compared to the image datasets in the computer vision community, the multi-modal datasets are still relatively small. However, the dataset size has grown by two orders of magnitudes between 2014 and 2019 (cf. Fig.~\ref{fig:dataset}(b)).

\subsection{Labels}
Most of the reviewed datasets provide ground-truth labels for 2D object detection and semantic segmentation tasks~\cite{hwang2015multispectral,choi2018kaist,Geiger2013IJRR,apolloscape_arXiv_2018,takumi2017multispectral,ha2017mfnet,braun2019eurocity}. KITTI~\cite{Geiger2013IJRR} also labels tracking, optical flow, visual odometry, and depth for various computer vision problems. BLV3D~\cite{blvdICRA2019} provides labels for tracking, interaction and intention. Labels for 3D scene understanding are provided by \cite{Geiger2013IJRR,apolloscape_arXiv_2018,360LiDARTracking_ICRA_2019, blvdICRA2019, nuscenes2019, lyft2019, Chang_2019_CVPR, pandaset, waymo_open_dataset}. 

Depending on the focus of a dataset, objects are labeled into different classes. For example, \cite{hwang2015multispectral} only contains label for people, including distinguishable individuals (labeled as ``Person"), non-distinguishable individuals (labeled as ``People"), and cyclists; \cite{apolloscape_arXiv_2018} classifies objects into five groups, and provides 25 fine-grained labels, such as truck, tricycle, traffic cone, and trash can. The Eurocity dataset \cite{braun2019eurocity} focuses on vulnerable road-users (mostly pedestrian). Instead of labeling objects, \cite{jung2016multi} provides a dataset for place categorization. Scenes are classified into forest, coast, residential area, urban area and indoor/outdoor parking lot. \cite{chen2018lidar} provides vehicle speed and wheel angles for driving behavior predictions. The BLV3D dataset~\cite{blvdICRA2019} provides unique labeling for interaction and intention.

\begin{figure}[tbp]
	\centering
		\centering
	\includegraphics[width=1.0\linewidth]{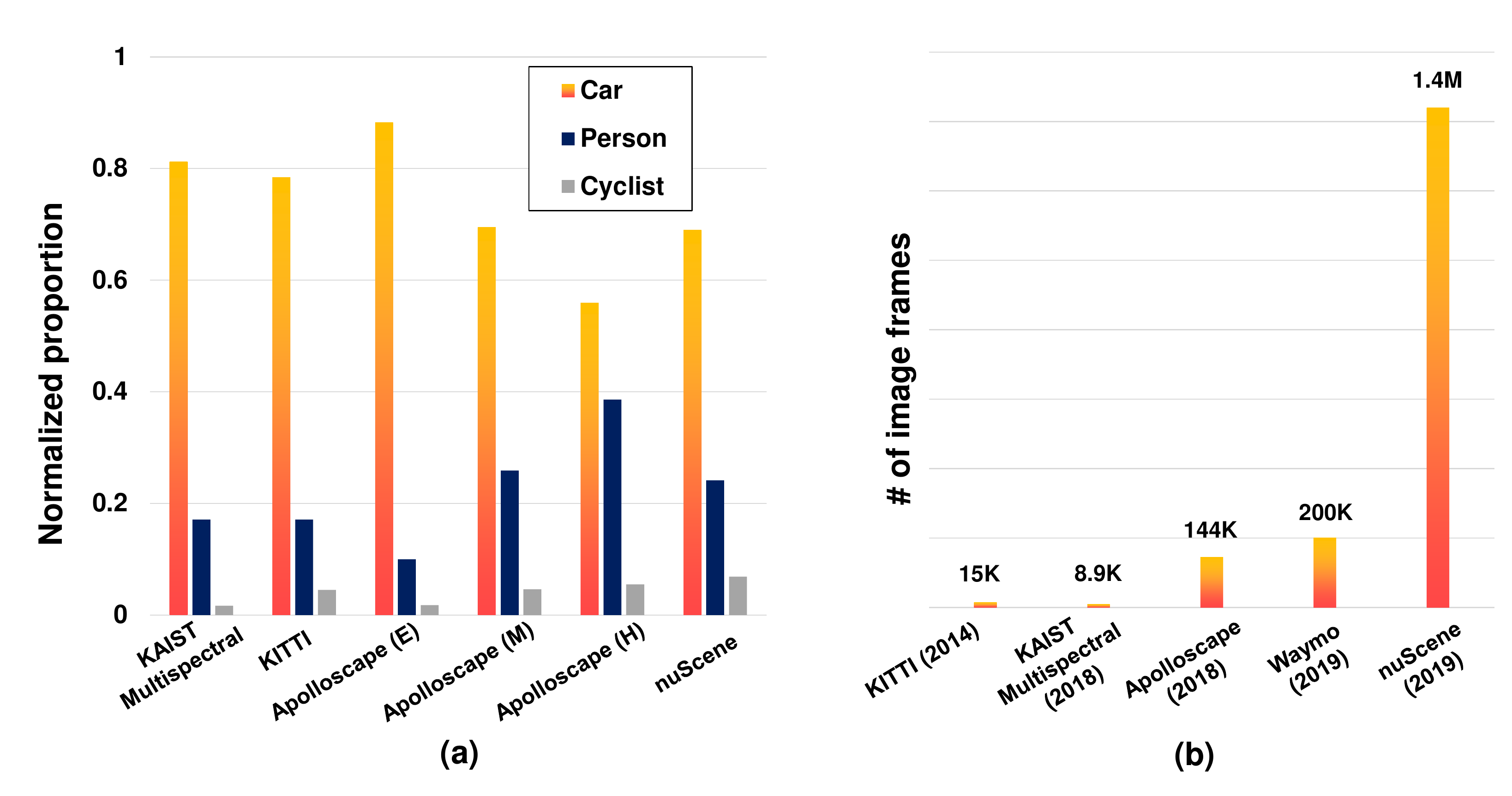}
	\caption{(a). Normalized percentage of objects of car, person, and cyclist classes in KAIST Multispectral~\cite{choi2018kaist}, KITTI~\cite{Geiger2012CVPR},  Apolloscape~\cite{apolloscape_arXiv_2018} (E: easy, M: moderate, and H: hard refer to the number of moveable objects in the frame - details can be found in~\cite{apolloscape_arXiv_2018}), and nuScene dataset~\cite{nuscenes2019}. (b). Number of camera image frames in several datasets. An increase by two orders of magnitude of the dataset size can be seen.}\label{fig:dataset}
\end{figure}

The object classes are very imbalanced. Fig.~\ref{fig:dataset}(a) compares the percentage of car, person, and cyclist classes from four reviewed datasets. There are much more objects labeled as car than person or cyclist.

\section{\textbf{Deep Multi-modal Perception Problems for Autonomous Driving}}\label{sec:perception}
In this section, we summarize deep multi-modal perception problems for autonomous driving based on sensing modalities and targets. An overview of the existing methods is shown in Tab.~\ref{tab:object_detection_summary} and Tab.~\ref{tab:object_segementation_summary}. An accuracy and runtime comparison among several methods is shown in Tab.~\ref{tab:perf_3d} and Tab.~\ref{tab:perf_semseg}.
\subsection{Deep Multi-modal Object Detection} \label{subsec:object_detection}
\subsubsection{Sensing Modalities} 
Most existing works combine RGB images from visual cameras with 3D LiDAR point clouds \cite{chen2017multi,asvadi2017multimodal,oh2017object,schlosser2016fusing,wang2017fusing,ku2017joint,xu2017pointfusion,qi2017frustum,du2017car,du2018general,matti2017combining,kim2016robust,kim2018robust,pfeuffer2018optimal,bijelic2019seeing, sindagi2019mvx, dou2019seg, wang2019frustum, liang2019multi}. Some other works focus on fusing the RGB images from visual cameras with images from thermal cameras \cite{wagner2016multispectral,liu2016bmvc,guan2018fusion,takumi2017multispectral}. Furthermore, Mees \textit{et al.}~\cite{mees2016choosing} employ a Kinect RGB-D camera to fuse RGB images and depth images; Schneider \textit{et al.}~\cite{schneider2017multimodal} generate depth images from a stereo camera and combine them with RGB images; Yang \textit{et al.}~\cite{yang2018hdnet} and Cascas \textit{et al.}~\cite{casas2018intentnet} leverage HD maps to provide prior knowledge of the road topology.

\subsubsection{2D or 3D Detection} 
Many works \cite{asvadi2017multimodal,oh2017object,du2017car,matti2017combining,kim2016robust,schneider2017multimodal,mees2016choosing,wagner2016multispectral,takumi2017multispectral,liu2016bmvc,schlosser2016fusing,guan2018fusion,kim2018season,pfeuffer2018optimal} deal with the 2D object detection problem on the front-view 2D image plane. Compared to 2D detection, 3D detection is more challenging since the object's distance to the ego-vehicle needs to be estimated. Therefore, accurate depth information provided by LiDAR sensors is highly beneficial. In this regard, some papers including \cite{chen2017multi,wang2017fusing,ku2017joint,xu2017pointfusion,qi2017frustum,du2018general,wang2019frustum,sindagi2019mvx} combine RGB camera images and LiDAR point clouds for 3D object detection. In addition, Liang \textit{et al.}~\cite{liang2019multi} propose a multi-task learning network to aid 3D object detection. The auxiliary tasks include camera depth completion, ground plane estimation, and 2D object detection. How to represent the modalities properly is discussed in section \ref{subsec:what_to_fuse}.

\subsubsection{What to detect}
Complex driving scenarios often contain different types of road users. Among them, cars, cyclists, and pedestrians are highly relevant to autonomous driving. In this regard, \cite{chen2017multi,asvadi2017multimodal,du2017car,matti2017combining,kim2018robust} employ multi-modal neural networks for car detection; \cite{schlosser2016fusing,matti2017combining,kim2016robust,mees2016choosing,wagner2016multispectral,liu2016bmvc,guan2018fusion} focus on detecting non-motorized road users (pedestrians or cyclists); \cite{oh2017object,wang2017fusing,ku2017joint,xu2017pointfusion,qi2017frustum,schneider2017multimodal,takumi2017multispectral,pfeuffer2018optimal,wang2019frustum,liang2019multi} detect both.

\subsection{Deep Multi-modal Semantic Segmentation} \label{subsec:segmentation}
Compared to the object detection problem summarized in Sec.~\ref{subsec:object_detection}, there are fewer works on multi-modal semantic segmentation: \cite{guan2018fusion, ha2017mfnet, sun2019rtfnet} employ RGB and thermal images, \cite{schneider2017multimodal} fuses RGB images and depth images from a stereo camera, \cite{valada2016deep,valada2017adapnet,valada2018self} combine RGB, thermal, and depth images for semantic segmentation in diverse environments such as forests, \cite{kim2018season} fuses RGB images and LiDAR point clouds for off-road terrain segmentation and \cite{yang2018fusion,caltagirone2019lidar,lv2018novel,wulff2018early, chen2019progressive} for road segmentation. Apart from the above-mentioned works for semantic segmentation on the 2D image plane,~\cite{piewak2018boosting,valada2016deep} deal with 3D segmentation on LiDAR points.

\section{\textbf{Methodology}}\label{sec:methodology}
When designing a deep neural network for multi-modal perception, three questions need to be addressed - \textit{What to fuse}: what sensing modalities should be fused, and how to represent and process them in an appropriate way; \textit{How to fuse}: what fusion operations should be utilized; \textit{When to fuse}: at which stage of feature representation in a neural network should the sensing modalities be combined. In this section, we summarize existing methodologies based on these three aspects.  
\subsection{What to Fuse}\label{subsec:what_to_fuse}
LiDARs and cameras (visual cameras, thermal cameras) are the most common sensors for multi-modal perception in the literature. While the interest in processing Radar signals via deep learning is growing, only a few papers discuss deep multi-modal perception with Radar for autonomous driving (e.g.~\cite{chadwick2019distant}). Therefore, we focus on several ways to represent and process LiDAR point clouds and camera images separately, and discuss how to combine them together. In addition, we briefly summarize Radar perception using deep learning. 

\subsubsection{LiDAR Point Clouds} 
LiDAR point clouds provide both depth and reflectance information of the environment. The depth information of a point can be encoded by its Cartesian coordinates $[x,y,z]$, distance $\sqrt{x^2+y^2+z^2}$, density, or HHA features (Horizontal disparity, Height, Angle)~\cite{gupta2014learning}, or any other 3D coordinate system. The reflectance information is given by intensity. 

There are mainly three ways to process point clouds. One way is by discretizing the 3D space into 3D voxels and assigning the points to the voxels (e.g.~\cite{zhou2017voxelnet,sindagi2019mvx,li20163d,engelcke2017vote3deep,shi2019part}). In this way, the rich 3D shape information of the driving environment can be preserved. However, this method results in many empty voxels as the LiDAR points are usually sparse and irregular. Processing the sparse data via clustering (e.g.~\cite{oh2017object,du2017car,du2018general,matti2017combining}) or 3D CNN (e.g.~\cite{li20163d,engelcke2017vote3deep}) is usually very time-consuming and infeasible for online autonomous driving. 
Zhou \textit{et al.}~\cite{zhou2017voxelnet} propose a voxel feature encoding (VFE) layer to process the LiDAR points efficiently for 3D object detection. They report an inference time of $\unit[225]{ms}$ on the KITTI dataset. Yan \textit{et al.}~\cite{yan2018second} add several sparse convolutional layers after the VFE to convert the sparse voxel data into 2D images, and then perform 3D object detection on them. Unlike the common convolution operation, the sparse convolution only computes on the locations associated with input points. In this way, they save a lot of computational cost, achieving an inference time of only $\unit[25]{ms}$.  

The second way is to directly learn over 3D LiDAR points in continuous vector space without voxelization. PointNet~\cite{qi2017pointnet} and its improved version PointNet++~\cite{qi2017pointnet++} propose to predict individual features for each point and aggregate the features from several points via max pooling. This method was firstly introduced in 3D object recognition and later extended by Qi \textit{et al.}~\cite{qi2017frustum}, Xu \textit{et al.}~\cite{xu2017pointfusion} and Shin \textit{et al.}~\cite{shin2018roarnet} to 3D object detection in combination with RGB images. Furthermore, Wang \textit{et al.}~\cite{wang2018deep} propose a new learnable operator called Parametric Continuous Convolution to aggregate points via a weighted sum, and Li \textit{et al.}~\cite{li2018pointcnn} propose to learn a $\chi$ transformation before applying transformed point cloud features into standard CNN. They are tested in semantic segmentation or LiDAR motion estimation tasks.

\begin{figure}[tbp]
	\centering
		\centering
	\includegraphics[width=0.96\linewidth]{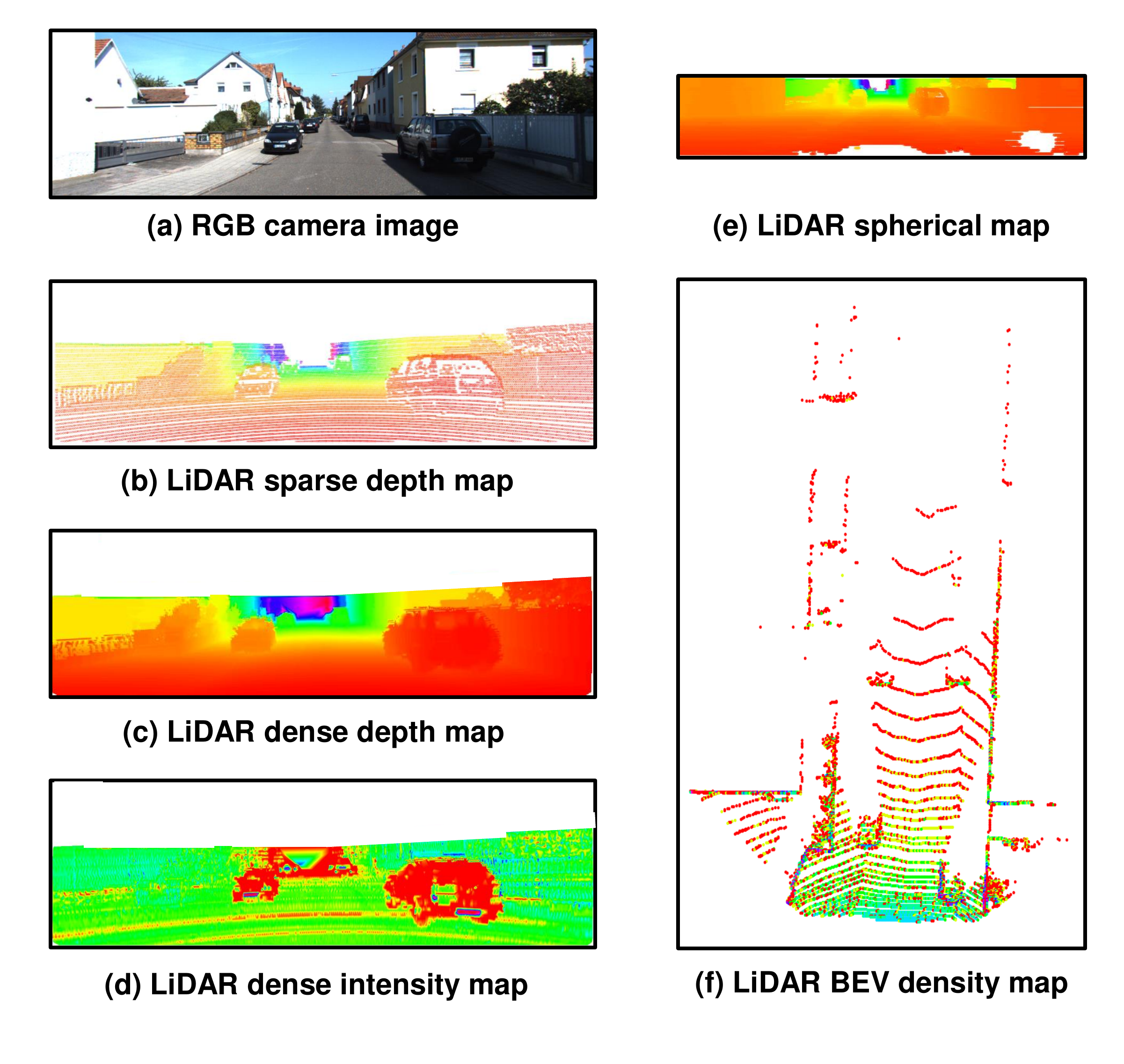}
	\caption{RGB image and different 2D LiDAR representation methods. (a) A standard RGB image, represented by a pixel grid and color channel values. (b) A sparse (front-view) depth map obtained from LiDAR measurements represented on a grid. (c) Interpolated depth map. (d) Interpolation of the measured reflectance values on a grid. (e) Interpolated representation of the measured LiDAR points (surround view) on a spherical map. (f)  Projection of the measured LiDAR points (front-facing) to bird's eye view (no interpolation).}\label{fig:lidar_images}
	\vspace{-0.3cm}
\end{figure}

A third way to represent 3D point clouds is by projecting them onto 2D grid-based feature maps so that they can be processed via 2D convolutional layers. In the following, we distinguish among spherical map, camera-plane map (CPM), as well as bird's eye view (BEV) map. Fig.~\ref{fig:lidar_images} illustrates different LiDAR representations in 2D. 

A spherical map is obtained by projecting each 3D point onto a sphere, characterized by azimuth and zenith angles. It has the advantage of representing each 3D point in a dense and compact way, making it a suitable representation for point cloud segmentation (e.g.~\cite{wu2017squeezeseg}). However, the size of the representation can be different from camera images. Therefore, it is difficult to fuse them at an early stage. 
A CPM can be produced by projecting the 3D points into the camera coordinate system, provided the calibration matrix. A CPM can be directly fused with camera images, as their sizes are the same. However, this representation leaves many pixels empty. Therefore, many methods have been proposed to up-sample such a sparse feature map, e.g. mean average~\cite{pfeuffer2018optimal}, nearest neighbors~\cite{asvadi2017depthcn}, or bilateral filter~\cite{cristinao2016}. 
Compared to the above-mentioned feature maps which encode LiDAR information in the front-view, a BEV map avoids occlusion problems because objects occupy different space in the map. In addition, the BEV preserves the objects' length and width, and directly provides the objects' positions on the ground plane, making the localization task easier. Therefore, the BEV map is widely applied to 3D environment perception. For example, Chen \textit{et al.}~\cite{chen2017multi} encode point clouds by height, density and intensity maps in BEV. The height maps are obtained by dividing the point clouds into several slices. The density maps are calculated as the number of points within a grid cell, normalized by the number of channels. The intensity maps directly represent the reflectance measured by the LiDAR on a grid. Lang \textit{et al.}~\cite{lang2018pointpillars} argue that the hard-coded features for BEV representation may not be optimal. They propose to learn features in each column of the LiDAR BEV representation via PointNet~\cite{qi2017pointnet}, and feed these learnable feature maps to standard 2D convolution layers. \\

\subsubsection{Camera Images}
Most methods in the literature employ RGB images from visual cameras or one type of infrared images from thermal cameras (near-infrared, mid-infrared, far-infrared). Besides, some works extract additional sensing information, such as optical flow~\cite{mees2016choosing}, depth~\cite{schneider2017multimodal,valada2017adapnet,valada2016deep}, or other multi-spectral images~\cite{valada2016deep,takumi2017multispectral}.

Camera images provide rich texture information of the driving surroundings. However, objects can be occluded and the scale of a single object can vary significantly in the camera image plane. For 3D environment inference, the bird's eye view that is commonly used for LiDAR point clouds might be a better representation. Roddick \textit{et al.}~\cite{roddick2018orthographic} propose a Orthographic Feature Transform (OFT) algorithm to project the RGB image features onto the BEV plane. The BEV feature maps are further processed for 3D object detection from monocular camera images. Lv \textit{et al.}~\cite{lv2018novel} project each image pixel with the corresponding LiDAR point onto the BEV plane and fuse the multi-modal features for road segmentation. Wang \textit{et al.}~\cite{wang2018pseudo} and their successive work~\cite{you2019pseudo} propose to convert RGB images into pseudo-lidar representation by estimating the image depth, and then use state-of-the-art BEV LiDAR detector to significantly improve the detection performance.

\subsubsection{Processing LiDAR Points and Camera Images in Deep Multi-modal Perception}~\label{subsubsec:process_lidar_camera} 
Tab.~\ref{tab:object_detection_summary} and Tab.~\ref{tab:object_segementation_summary} summarize existing methods to process sensors' signals for deep multi-modal perception, mainly LiDAR points and camera images. From the tables we have three observations: (1). Most works propose to fuse LiDAR and camera features extracted from 2D convolution neural networks. To do this, they project LiDAR points on the 2D plane and process the feature maps through 2D convolutions. Only a few works extract LiDAR features by PointNet (e.g.~\cite{xu2017pointfusion,qi2017frustum,yang2018fusion}) or 3D convolutions (e.g.~\cite{kim2018season}); (2). Several works on multi-modal object detection cluster and segment 3D LiDAR points to generate 3D region proposals (e.g.~\cite{oh2017object,du2017car,matti2017combining}). Still, they use a LiDAR 2D representation to extract features for fusion; (3). Several works project LiDAR points on the camera-plane or RGB camera images on the LiDAR BEV plane (e.g.~\cite{lv2018novel,wulff2018early,liang2018deep}) in order to align the features from different sensors, whereas many works propose to fuse LiDAR BEV features directly with RGB camera images (e.g.~\cite{chen2017multi,ku2017joint}). This indicates that the networks implicitly learn to align features of different viewpoints. Therefore, a well-calibrated sensor setup with accurate spatial and temporal alignment is the prerequisite for accurate multi-modal perception, as will be discussed in Sec.~\ref{subsubsec:data_quality}.

\subsubsection{Radar Signals}
Radars provide rich environment information based on received amplitudes, ranges, and the Doppler spectrum. The Radar data can be represented by 2D feature maps and processed by convolutional neural networks. For example, Lombacher \textit{et al.} employ Radar grid maps made by accumulating Radar data over several time-stamps~\cite{werber2015automotive} for static object classification~\cite{lombacher2016potential} and semantic segmentation ~\cite{lombacher2017semantic} in autonomous driving. Visentin \textit{et al.} show that CNNs can be employed for object classification in a post-processed range-velocity map \cite{visentin2017classification}. Kim \textit{et al.}~\cite{kim2018moving} use a series of Radar range-velocity images and convolutional recurrent neural networks for moving objects classification. Moeness \textit{et al.}~\cite{amin2018understanding} feed spectrogram from Time Frequency signals as 2D images into a stacked auto-encoders to extract high-level Radar features for human motion recognition. The Radar data can also be represented directly as ``point clouds" and processed by PointNet++~\cite{qi2017pointnet++} for dynamic object segmentation~\cite{schumann2018semantic}. Besides, Woehler \textit{et al.}~\cite{wohler2017comparison} encode features from a cluster of Radar points for dynamic object classification. Chadwick \textit{et al.}~\cite{chadwick2019distant} first project Radar points on the camera plane to build Radar range-velocity images, and then combine with camera images for distant vehicle detection.

\subsection{How to Fuse} \label{subsec:how_to_fuse}
This section summarizes typical fusion operations in a deep neural network. For simplicity we restrict our discussion to two sensing modalities, though more still apply. Denote $M_i$ and $M_j$ as two different modalities, and $f_l^{M_i}$ and $f_l^{M_j}$ their feature maps in the $l^{th}$ layer of the neural network. Also denote $G_l(\cdot)$ as a mathematical description of the feature transformation applied in layer $l$ of the neural network. 

\subsubsection{Addition or Average Mean}
This join operation adds the feature maps element-wise, i.e. $f_l = G_{l-1}\left(f_{l-1}^{M_i}+f_{l-1}^{M_j}\right)$, or calculates the average mean of the feature maps.

\subsubsection{Concatenation}
Combines feature maps by $f_l = G_{l-1}\left(f_{l-1}^{M_i\ \frown} f_{l-1}^{M_j} \right)$. The feature maps are usually stacked along their depth before they are advanced to a convolution layer. For a fully connected layer, these features are usually flattened into vectors and concatenated along the rows of the feature maps.

\subsubsection{Ensemble}
This operation ensembles feature maps from different sensing modalities via $f_l = G_{l-1}\left(f_{l-1}^{M_i}\right) \cup G_{l-1}\left(f_{l-1}^{M_j}\right)$. As will be introduced in the following sections (Sec.~\ref{subsubsc:method:fusion_detection} and Sec.~\ref{subsubsc:method:operation_scheme}), ensembles are often used to fuse ROIs in object detection networks.

\subsubsection{Mixture of Experts}
\begin{figure}[tbp]
	\centering
		\centering
	\includegraphics[width=0.88\linewidth]{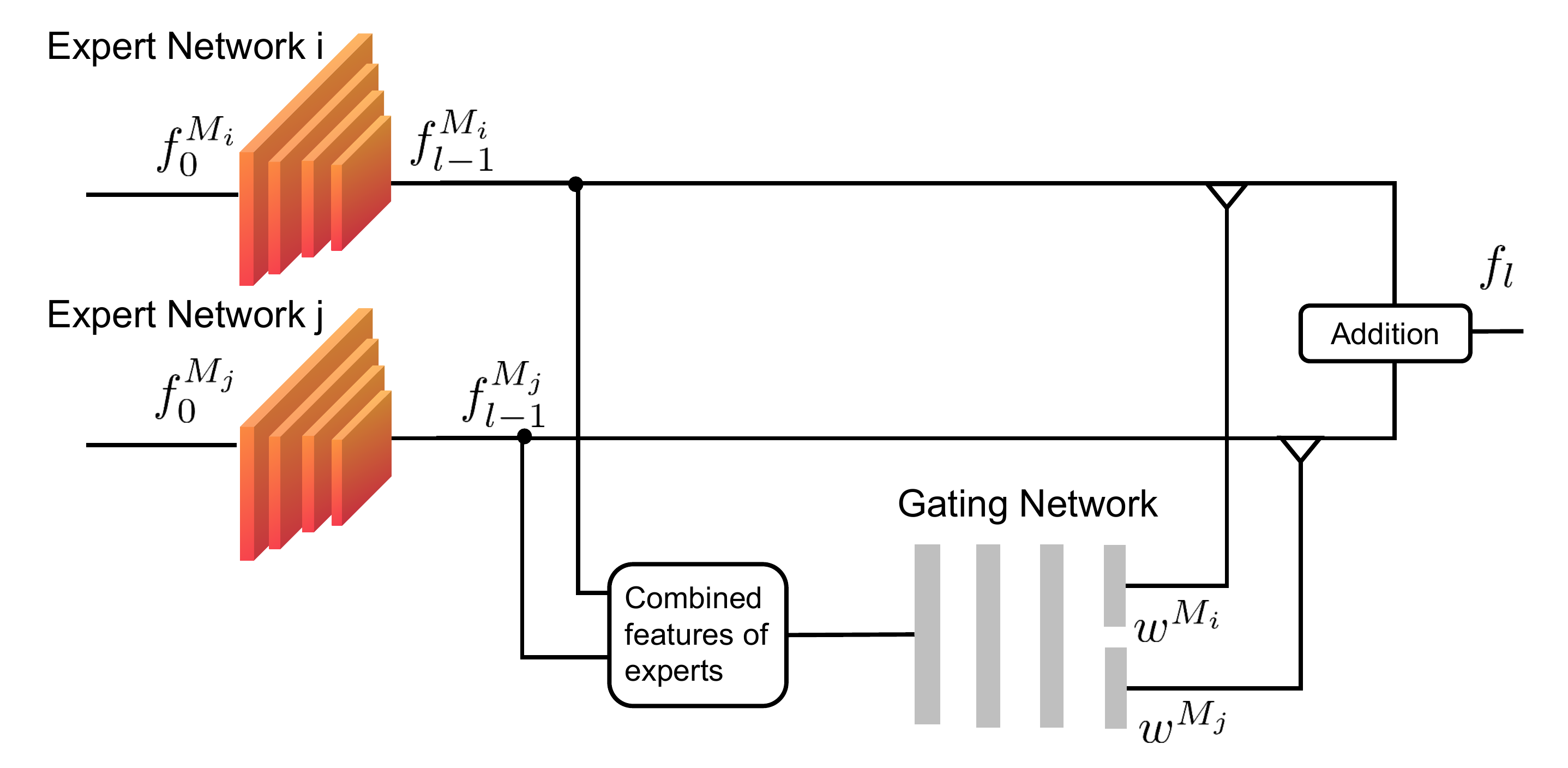}
	\caption{An illustration of the Mixture of Experts fusion method. Here we show the combined features which are derived from the output layers of the expert networks. They can be extracted from the intermediate layers as well.}\label{fig:mixture_of_experts}
\end{figure}

The above-mentioned fusion operations do not consider the informativeness of a sensing modality (e.g. at night time RGB camera images bring less information than LiDAR points). These operations are applied, hoping that the network can \textit{implicitly} learn to weight the feature maps. In contrast, the Mixture of Experts (MoE) approach \textit{explicitly} models the weight of a feature map. It is first introduced in~\cite{Jacobs1991AdaptiveMO} for neural networks and then extended in~\cite{eigen2014,mees2016choosing,valada2017adapnet}. As Fig.~\ref{fig:mixture_of_experts} illustrates, the feature map of a sensing modality is processed by its domain-specific network called     ``expert". Afterwards, the outputs of multiple expert networks are averaged with the weights $w^{M_i}, w^{M_j}$ predicted by a gating network which takes the combined features output by the expert networks as inputs $h$ via a simple fusion operation such as concatenation:
\begin{equation}
f_l = G_l\left(w^{M_i}\cdot f^{M_i}_{l-1} + w^{M_j}\cdot f^{M_j}_{l-1}\right),\ \text{with}\ w^{M_i}+w^{M_j}=1. 
\end{equation}

\begin{figure*}[tbp]
	\centering
	\includegraphics[width=0.86\textwidth]{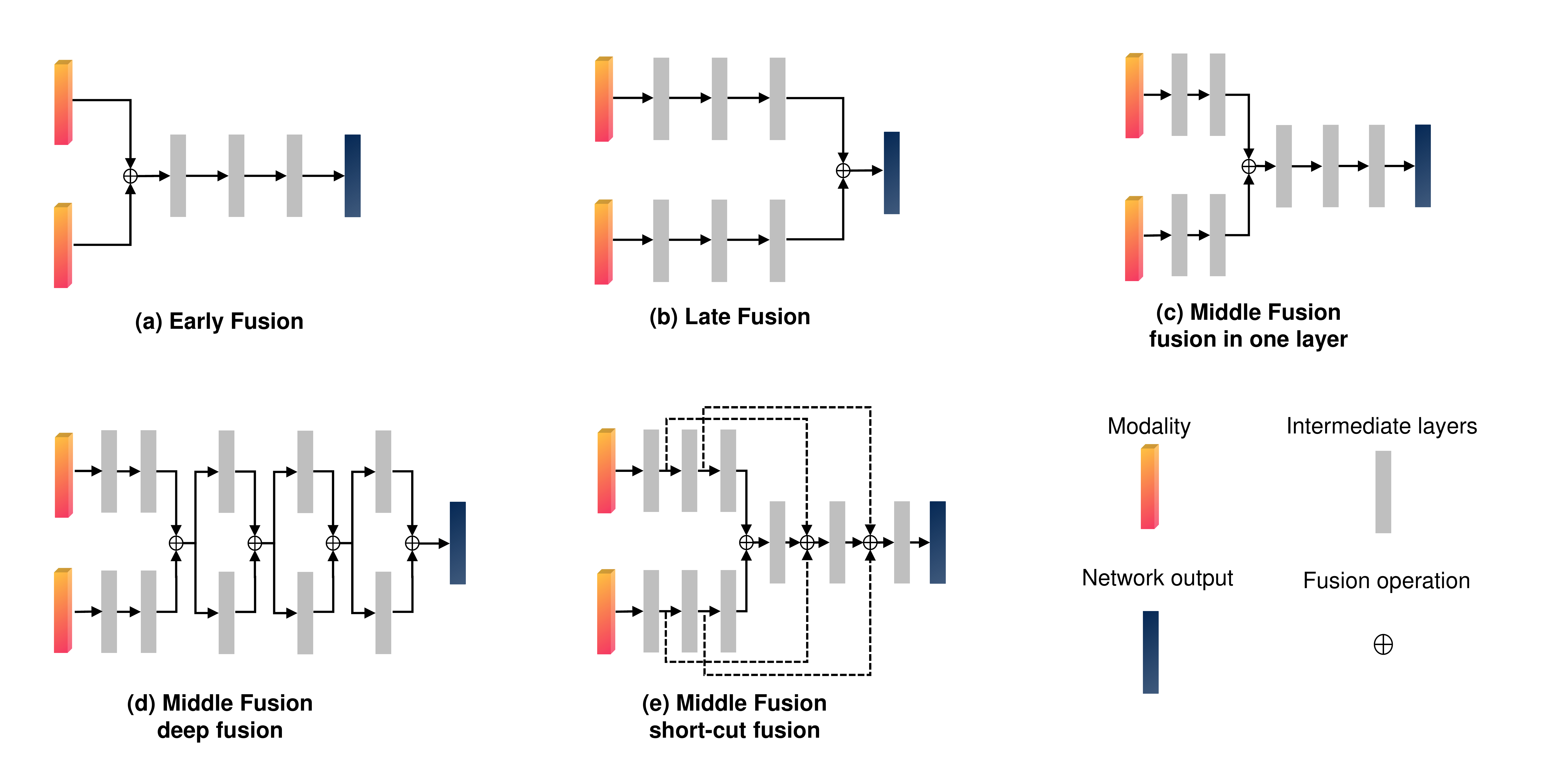}
	\caption{An illustration of early fusion, late fusion, and several middle fusion methods.}\label{fig:fusion_network_architecture}
    \centering
\end{figure*}

\subsection{When to Fuse} \label{subsec:when_to_fuse}
Deep neural networks represent features hierarchically and offer a wide range of choices to combine sensing modalities at early, middle, or late stages (Fig.~\ref{fig:fusion_network_architecture}). In the sequel, we discuss the early, middle, and late fusions in detail. For each fusion scheme, we first give mathematical descriptions using the same notations as in Sec.~\ref{subsec:how_to_fuse}, and then discuss their properties. Note that there exists some works that fuse features from the early stage till late stages in deep neural networks (e.g.~\cite{bloesch2018codeslam}). For simplicity, we categorize this fusion scheme as ``middle fusion".  Compared to the semantic segmentation where multi-modal features are fused at different stages in FCN, there exist more diverse network architectures and more fusion variants in object detection. Therefore, we additionally summarize the fusion methods specifically for the object detection problem. Finally, we discuss the relationship between the fusion operation and the fusion scheme. 

Note that we do not find conclusive evidence from the methods we have reviewed that one fusion method is better than the others. The performance is highly dependent on sensing modalities, data, and network architectures. 

\subsubsection{Early Fusion}
This method fuses the raw or pre-processed sensor data. Let us define $f_{l} = f_{l-1}^{M_i} \oplus f_{l-1}^{M_j}$ as a fusion operation introduced in Sec.~\ref{subsec:how_to_fuse}. For a network that has $L+1$ layers, an early fusion scheme can be described as:
\begin{equation}
f_L = G_L \Bigg(G_{L-1}\bigg(\cdots G_l\Big(\cdots G_2 \big(G_1(f_0^{M_i} \oplus f_0^{M_j})\big)\Big)\bigg)\Bigg),
\end{equation}
with $l=[1,2,\cdots,L]$. Early fusion has several pros and cons. First, the network learns the joint features of multiple modalities at an early stage, fully exploiting the information of the raw data. Second, early fusion has low computation requirements and a low memory budget as it jointly processes the multiple sensing modalities. This comes with the cost of model inflexibility. As an example, when an input is replaced with a new sensing modality or the input channels are extended, the early fused network needs to be retrained completely. Third, early fusion is sensitive to spatial-temporal data misalignment among sensors which are caused by calibration error, different sampling rate, and sensor defect.

\subsubsection{Late Fusion}
This fusion scheme combines decision outputs of each domain specific network of a sensing modality. It can be described as:
\begin{equation}
f_L = G^{M_i}_L\Big(G^{M_i}_{L-1}\big(\cdots G^{M_i}_1(f_0^{M_i})\big)\Big) \oplus G^{M_j}_L\Big(G^{M_j}_{L-1}\big(\cdots G^{M_j}_1(f_0^{M_j})\big)\Big).
\end{equation}
Late fusion has high flexibility and modularity. When a new sensing modality is introduced, only its domain specific network needs to be trained, without affecting other networks. However, it suffers from high computation cost and memory requirements. In addition, it discards rich intermediate features which may be highly beneficial when being fused.

\subsubsection{Middle Fusion}
Middle fusion is the compromise of early and late fusion: It combines the feature representations from different sensing modalities at intermediate layers. This enables the network to learn cross modalities with different feature representations and at different depths. Define $l^{\star}$ as the layer from which intermediate features begin to be fused. The middle fusion can be executed at this layer only once:
\begin{equation}
f_L = G_L\bigg(\cdots G_{l^{\star}+1}\Big(G^{M_i}_{l^{\star}}\big(\cdots G^{M_i}_1(f_0^{M_i})\big) \oplus G^{M_j}_{l^{\star}}\big(\cdots G^{M_j}_1(f_0^{M_j})\big)\Big)\bigg).
\end{equation}
Alternatively, they can be fused hierarchically, such as by deep fusion~\cite{WangWZZ16,chen2017multi}:
\begin{equation}
\begin{split}
&f_{l^{\star}+1} = f_{l^{\star}}^{M_i} \oplus f_{l^{\star}}^{M_j}, \\
&f_{k+1} = G^{M_i}_{k}(f_{k}) \oplus G^{M_j}_{k}(f_{k}),\ \forall  k: k\in \left\{l^{\star}+1,\cdots,L\right\}.
\end{split}
\end{equation}
or ``short-cut fusion''~\cite{ha2017mfnet}:
\begin{equation}
\begin{split}
&f_{l+1} = f_{l}^{M_i} \oplus f_{l}^{M_j}, \\
&f_{k+1} = f_{k} \oplus f^{M_i}_{k^\star} \oplus  
f^{M_j}_{k^\star}, \\
& \forall k: k\in \left\{l+1,\cdots,L\right\}; \ \exists k^\star: k^\star \in \left\{1,\cdots,l-1\right\}.
\end{split}
\end{equation}
Although the middle fusion approach is highly flexible, it is not easy to find the ``optimal'' way to fuse intermediate layers given a specific network architecture. We will discuss this challenge in detail in Sec.~\ref{subsubsec:challenge:when_to_fuse}.

\begin{figure*}[htbp]
	\centering
	\includegraphics[width=0.96\textwidth]{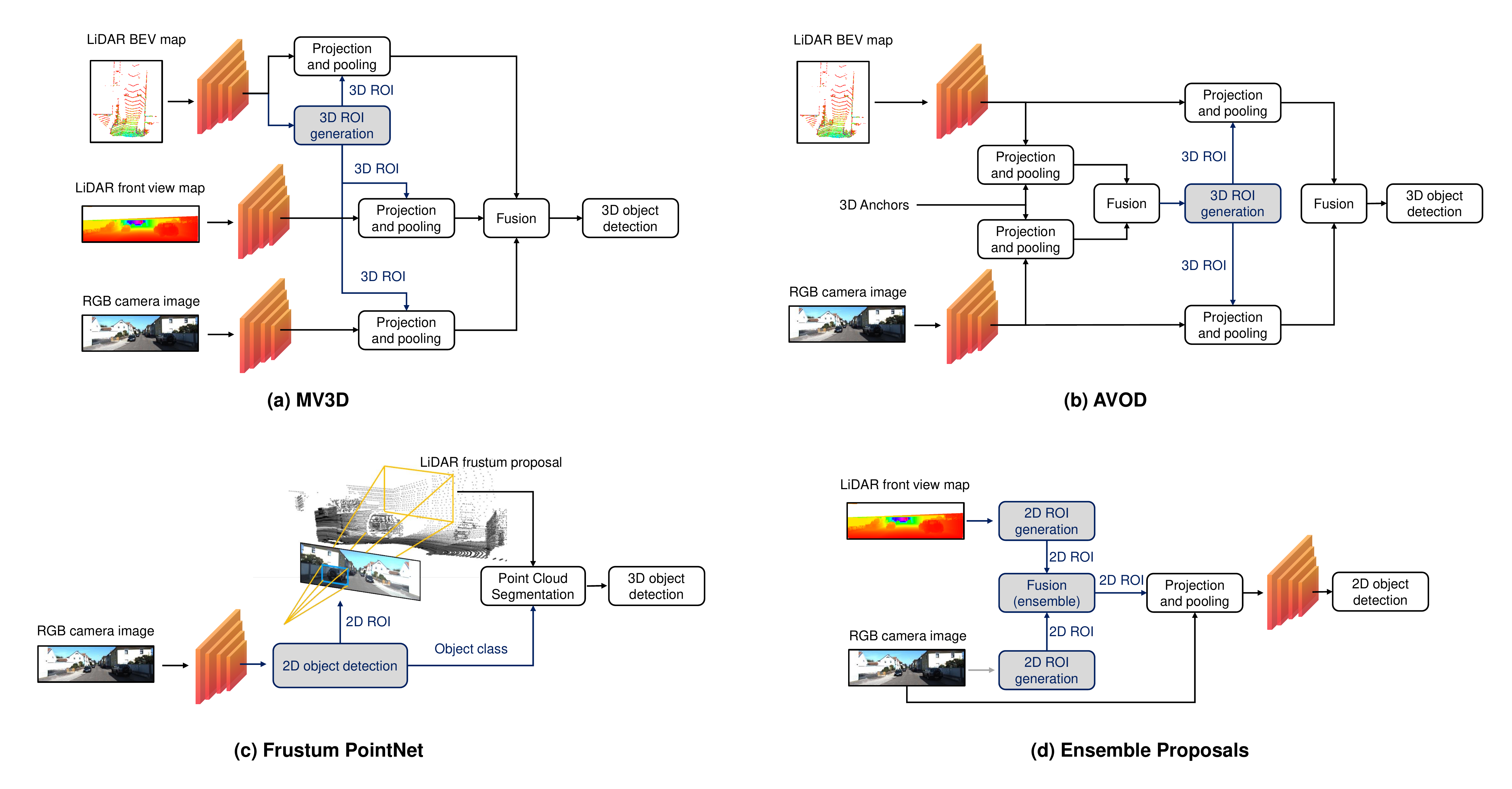}
	\caption{Examplary fusion architectures for two-stage object detection networks. (a). MV3D~\cite{chen2017multi}; (b). AVOD~\cite{ku2017joint}; (c). Frustum PointNet~\cite{qi2017frustum}; (d). Ensemble Proposals~\cite{kim2016robust}.   }\label{fig:2_stage_object_detection}
	\centering
\end{figure*}

\subsubsection{Fusion in Object Detection Networks}\label{subsubsc:method:fusion_detection}
Modern multi-modal object detection networks usually follow either the two-stage pipeline (RCNN~\cite{girshick2014rich}, Fast-RCNN~\cite{girshick2015fast}, Faster-RCNN~\cite{ren2015faster}) or the one-stage pipeline (YOLO~\cite{redmon2016you} and SSD~\cite{liu2016ssd}), as explained in detail in Sec.~\ref{subsec:deep_object_detection}. This offers a variety of alternatives for network fusion. For instance, the sensing modalities can be fused to generate regional proposals for a two-stage object detector. The regional multi-modal features for each proposal can be fused as well. Ku \textit{et al.}~\cite{ku2017joint} propose AVOD, an object detection network that fuses RGB images and LiDAR BEV images both in the region proposal network and the header network. Kim \textit{et al.}~\cite{kim2016robust} ensemble the region proposals that are produced by LiDAR depth images and RGB images separately. The joint region proposals are then fed to a convolutional network for final object detection. Chen \textit{et al.}~\cite{chen2017multi} use LiDAR BEV maps to generate region proposals. For each ROI, the regional features from the LiDAR BEV maps are fused with those from the LiDAR front-view maps as well as camera images via deep fusion. Compared to object detections from LiDAR point clouds, camera images have been well investigated with larger labeled dataset and better 2D detection performance. Therefore, it is straightforward to exploit the predictions from well-trained image detectors when doing camera-LiDAR fusion. In this regard,~\cite{qi2017frustum,du2018general,xu2017pointfusion} propose to utilize a pre-trained image detector to produce 2D bounding boxes, which build frustums in LiDAR point clouds. Then, they use these point clouds within the frustums for 3D object detection.
Fig.~\ref{fig:2_stage_object_detection} shows some exemplary fusion architectures for two-stage object detection networks.
Tab.~\ref{tab:object_detection_summary} summarizes the methodologies for multi-modal object detection. 

\subsubsection{Fusion Operation and Fusion Scheme}\label{subsubsc:method:operation_scheme}
Based on the papers that we have reviewed, feature concatenation is the most common operation, especially at early and middle stages. Element-wise average mean and addition operations are additionally used for middle fusion. Ensemble and Mixture of Experts are often used for middle to decision level fusion. 


\begin{table*}[tbp] 
	\centering
	\caption{AN OVERVIEW OF CHALLENGES AND OPEN QUESTIONS}\label{tab:challenges}
	\resizebox{1\textwidth}{!}{\begin{tabular}{c  c  c c}
			\rowcolor{lightgray!75}  
			\multicolumn{2}{c}{ \textbf{Topics}} &  \textbf{Challenges} & \textbf{Open Questions} \\
			
			\multirow{2}{*}{\parbox{0.2\linewidth}{\vspace{1.5cm} \ \ \ \  Multi-modal data preparation}}
			& Data diversity & 
			\begin{minipage}[t]{0.4\textwidth}  
				\begin{itemize} 
					\item Relative small size of training dataset. 
					\item Limited driving scenarios and conditions, limited sensor variety, object class imbalance.  
			\end{itemize} \end{minipage} & 
			
			\begin{minipage}[t]{0.4\textwidth}
				\begin{itemize} 
					\item Develop more realistic virtual datasets. 
					\item Finding optimal way to combine real- and virtual data.
					\item Increasing labeling efficiency through cross-modal labeling, active learning, transfer learning, semi-supervised learning etc. Leveraging lifelong learning to update networks with continual data collection. 
				\end{itemize} \vspace{+3mm} \end{minipage}  \\ 
			
			& Data quality & \begin{minipage}[t]{0.4\textwidth}  \begin{itemize} 
					\item Labeling errors. 
					\item Spatial and temporal misalignment of different sensors.
			\end{itemize} \end{minipage} & \begin{minipage}[t]{0.4\textwidth}
				\begin{itemize}
					\item Teaching network robustness with erroneous and noisy labels.
					\item Integrating prior knowledge in networks.
					\item  Developing methods (e.g. using deep learning) to automatically register sensors. \\
			\end{itemize} \end{minipage} \\ \hline
			
			\multirow{3}{*}{\parbox{0.2\linewidth}{\vspace{2.5cm} \ \ \ \ \ \ Fusion methodology}} &  ``What to fuse'' & 
			\begin{minipage}[t]{0.4\textwidth}  \begin{itemize} 
					\item Too few sensing modalities are fused.
					\item  Lack of studies for different feature representations.\\
				\end{itemize} 
			\end{minipage} & 
			
			\begin{minipage}[t]{0.4\textwidth}  \begin{itemize} 
					\item Fusing multiple sensors with the same modality.
					\item Fusing more sensing modalities, e.g. Radar, Ultrasonic, V2X communication.
					\item Fusing with physical models and prior knowledge, also possible in the multi-task learning scheme.
					\item Comparing different feature representation w.r.t informativeness and computational costs.
				\end{itemize} 
				\vspace{+3mm} \end{minipage}  \\  
			
			&``How to fuse'' & \begin{minipage}[t]{0.4\textwidth} \begin{itemize} 
					\item Lack of uncertainty quantification for each sensor channel.
					\item Too simple fusion operations.\\
			\end{itemize} \end{minipage}  & \begin{minipage}[t]{0.4\textwidth} \begin{itemize} 
					\item Uncertainty estimation via e.g. Bayesian neural networks (BNN).
					\item Propagating uncertainties to other modules, such as tracking and motion planning.
					\item Anomaly detection by generative models.
					\item Developing fusion operations that are suitable for network pruning and compression.\\
				\end{itemize}  \vspace{+3mm} \end{minipage} \\ 
			
			&``When to fuse'' & \begin{minipage}[t]{0.4\textwidth} \begin{itemize} 
					\item Fusion architecture is often designed by empirical results. No guideline for optimal fusion architecture design.
					\item Lack of study for accuracy/speed  or memory/robustness trade-offs.\\
			\end{itemize} \end{minipage} & \begin{minipage}[t]{0.4\textwidth} \begin{itemize} 
					\item Optimal fusion architecture search.
					\item Incorporating requirements of computation time or memory as regularization term.
					\item Using visual analytics tool to find optimal fusion architecture.\\
			\end{itemize} \end{minipage} \\ \hline
			
			\multirow{2}{*}{Others} & Evaluation metrics & \begin{minipage}[t]{0.4\textwidth} \begin{itemize} 
					\item Current metrics focus on comparing networks' accuracy.
			\end{itemize} \end{minipage} & \begin{minipage}[t]{0.4\textwidth} \begin{itemize} 
					\item Metrics to quantify the networks' robustness should be developed and adapted to multi-modal perception problems. \\
				\end{itemize}  \vspace{+3mm} \end{minipage} \\ 
			
			& More network architectures & \begin{minipage}[t]{0.4\textwidth} \begin{itemize} 
					\item Current networks lack temporal cues and cannot guarantee prediction consistency over time. 
					\item They are designed mainly for modular autonomous driving.\\
			\end{itemize} \end{minipage}  & \begin{minipage}[t]{0.4\textwidth} \begin{itemize} 
					\item Using Recurrent Neural Network (RNN) for sequential perception. 
					\item Multi-modal end-to-end learning or multi-modal direct-perception.\\
			\end{itemize} \end{minipage} \\ \bottomrule
	\end{tabular}}
\end{table*}

\section{\textbf{Challenges and Open Questions}}\label{sec:challenge}
As discussed in the Introduction (cf. Sec.~\ref{sec:intro}), developing deep multi-modal perception systems is especially challenging for autonomous driving because it has high requirements in accuracy, robustness, and real-time performance. The predictions from object detection or semantic segmentation are usually transferred to other modules such as maneuver prediction and decision making. A reliable perception system is the prerequisite for a driverless car to run safely in uncontrolled and complex driving environments. In Sec.~\ref{sec:dataset} and Sec.~\ref{sec:methodology} we have summarized the multi-modal datasets and fusion methodologies. Correspondingly, in this section we discuss the remaining challenges and open questions for multi-modal data preparation and network architecture design. We focus on how to improve the accuracy and robustness of the multi-modal perception systems while guaranteeing real-time performance. We also discuss some open questions, such as evaluation metrics and network architecture design. Tab.~\ref{tab:challenges} summarizes the challenges and open questions.

\subsection{Multi-modal Data Preparation}
\subsubsection{Data Diversity}\label{subsubsec:data_diversity}
Training a deep neural network on a complex task requires a huge amount of data. Therefore, using large multi-modal datasets with diverse driving conditions, object labels, and sensors can significantly improve the network's accuracy and robustness against changing environments. However, it is not an easy task to acquire real-world data due to cost and time limitations as well as hardware constraints. The size of open multi-modal datasets is usually much smaller than the size of image datasets. As a comparison, KITTI~\cite{Geiger2012CVPR} records only 80,256 objects whereas ImageNet~\cite{imagenet} provides 1,034,908 samples. Furthermore, the datasets are usually recorded in limited driving scenarios, weather conditions, and sensor setups (more details are provided in Sec.~\ref{sec:dataset}). The distribution of objects is also very imbalanced, with much more objects being labeled as car than person or cyclist (Fig.~\ref{fig:dataset}). As a result, it is questionable how a deep multi-modal perception system trained with those public datasets performs when it is deployed to an unstructured environment.

One way to overcome those limitations is by data augmentation via simulation. In fact, a recent work~\cite{ngiam2019starnet} states that the most performance gain for object detection in the KITTI dataset is due to data augmentation, rather than advances in network architectures. Pfeuffer \textit{et al.}~\cite{pfeuffer2018optimal} and Kim \textit{et al.}~\cite{kim2018robust} build augmented training datasets by adding artificial blank areas, illumination change, occlusion, random noises, etc. to the KITTI dataset. The datasets are used to simulate various driving environment changes and sensor degradation. They show that trained with such datasets, the network accuracy and robustness are improved. Some other works aim at developing virtual simulators to generate varying driving conditions, especially some dangerous scenarios where collecting real-world data is very costly or hardly possible. Gaidon \textit{et al.}~\cite{Gaidon:Virtual:CVPR2016} build a virtual KITTI dataset by introducing a real to virtual cloning method to the original KITTI dataset, using the Unity Game Engine. Other works~\cite{Richter_2016_ECCV,ros2016synthia,richter2017playing,yue2018lidar,wrenninge2018synscapes,DeepMVS} generate virtual datasets purely from game engines, such as GTA-V, without a proxy of real-world datasets. Griffiths and Boehm~\cite{griffiths2019synthcity} create a purely virtual LiDAR only dataset. In addition, Dosovitskiy \textit{et al.}~\cite{Dosovitskiy17} develop an open-source simulator that can simulate multiple sensors in autonomous driving and Hurl \textit{et al.}~\cite{hurl2019precise} release a large scale, virtual, multi-modal dataset with LiDAR data and visual camera. Despite many available virtual datasets, it is an open question to which extend a simulator can represent real-world phenomena. Developing more realistic simulators and finding the optimal way to combine real and virtual data are important open questions.

\begin{figure}[tbp]
	\centering
	\centering
	\includegraphics[width=0.98\linewidth]{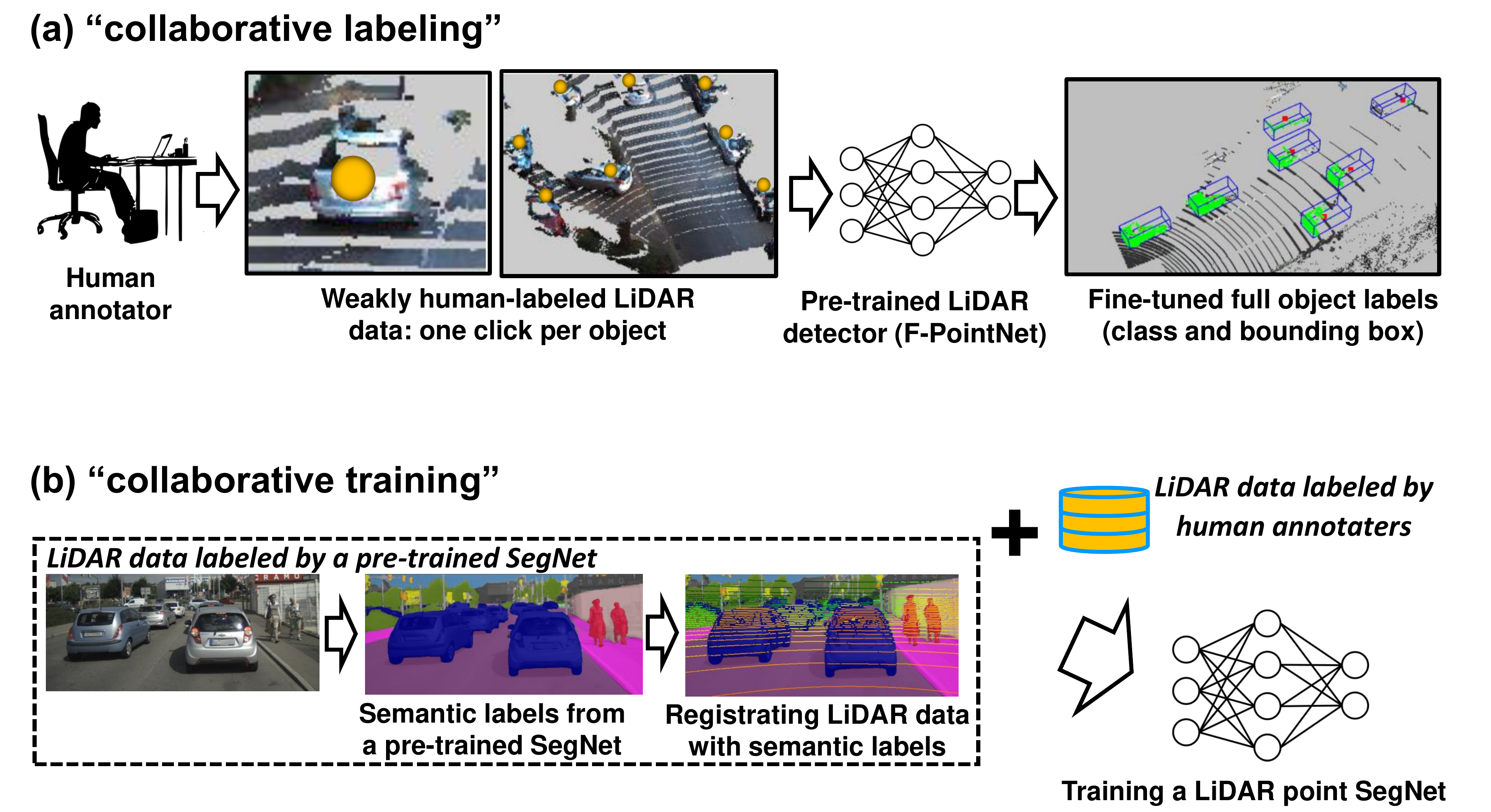}
	\caption{Two examples of increasing data labeling efficiency in LiDAR data. (a) Collaborative labeling LiDAR points for 3D detection~\cite{lee2018leveraging}: the LiDAR points within each object are firstly weakly-labeled by human annotators, and then fine-tuned by a pre-trained LiDAR detector based on the F-PointNet. (b) Collaborative training a semantic segmentation network  (SegNet) for LiDAR points~\cite{piewak2018boosting}: To boost the training data, a pre-trained image SegNet can be employed to transfer the image semantics. }\label{fig:cross_labeling}
	\vspace{-0.3cm}
\end{figure}
 
Another way to overcome the limitations of open datasets is by increasing the efficiency of data labeling. When building a multi-modal training dataset, it is relatively easy to drive the test vehicle and collect many data samples. However, it is very tedious and time-consuming to label them, especially when dealing with 3D labeling and LiDAR points. Lee \textit{et al.}~\cite{lee2018leveraging} develop a collaborative hybrid labeling tool, where 3D LiDAR point clouds are firstly weakly-labeled by human annotators, and then fine-tuned by pre-trained network based on F-PointNet~\cite{qi2017frustum}. They report that the labeling tool can significantly reduce the ``task complexity'' and ``task switching'', and have a 30$\times$ labeling speed-up (Fig.~\ref{fig:cross_labeling}(a)). Piewak \textit{et al.}~\cite{piewak2018boosting} leverage a pre-trained image segmentation network to label LiDAR point clouds without human intervention. The method works by registering each LiDAR point with an image pixel, and transferring the image semantics predicted by the pre-trained network to the corresponding LiDAR points (cf. Fig.~\ref{fig:cross_labeling}(b)). In another work, Mei \textit{et al.}~\cite{mei2018semantic} propose a semi-supervised learning method to do 3D point segmentation labeling. With only a few manual labels together with pair-wise spatial constraints between adjacent data frames, a lot of objects can be labeled. Several works \cite{mackowiak2018cereals, RoyUN18, feng2019deep} propose to introduce active learning in semantic segmentation or object detection for autonomous driving. The networks iteratively query the human annotator some most informative samples in an unlabeled data pool and then update the networks' weights. In this way, much less labeled training data is required while reaching the same performance and saving human labeling efforts. There are many other methods in the machine learning literature that aim to reduce data labeling efforts, such as transfer learning~\cite{pan2010survey}, domain adaptation~\cite{patel2015visual,chen2019learning,chen2018domain,lee2018spigan,tremblay2018training}, and semi-supervised learning~\cite{kingma2014semi}. How to efficiently label multi-modal data in autonomous driving is an important and challenging future work, especially in scenarios where the signals from different sensors may not be matched (e.g. due to the distance some objects are only visible by visual camera but not by LiDAR). Finally, as there can always be new driving scenarios that are different from the training data, it is an interesting research topic to leverage lifelong learning~\cite{parisi2019continual} to update the multi-modal perception network with continual data collection.

\subsubsection{Data Quality and Alignment}\label{subsubsec:data_quality}
\begin{figure}[tbp]
	\centering
		\centering
	\includegraphics[width=0.98\linewidth]{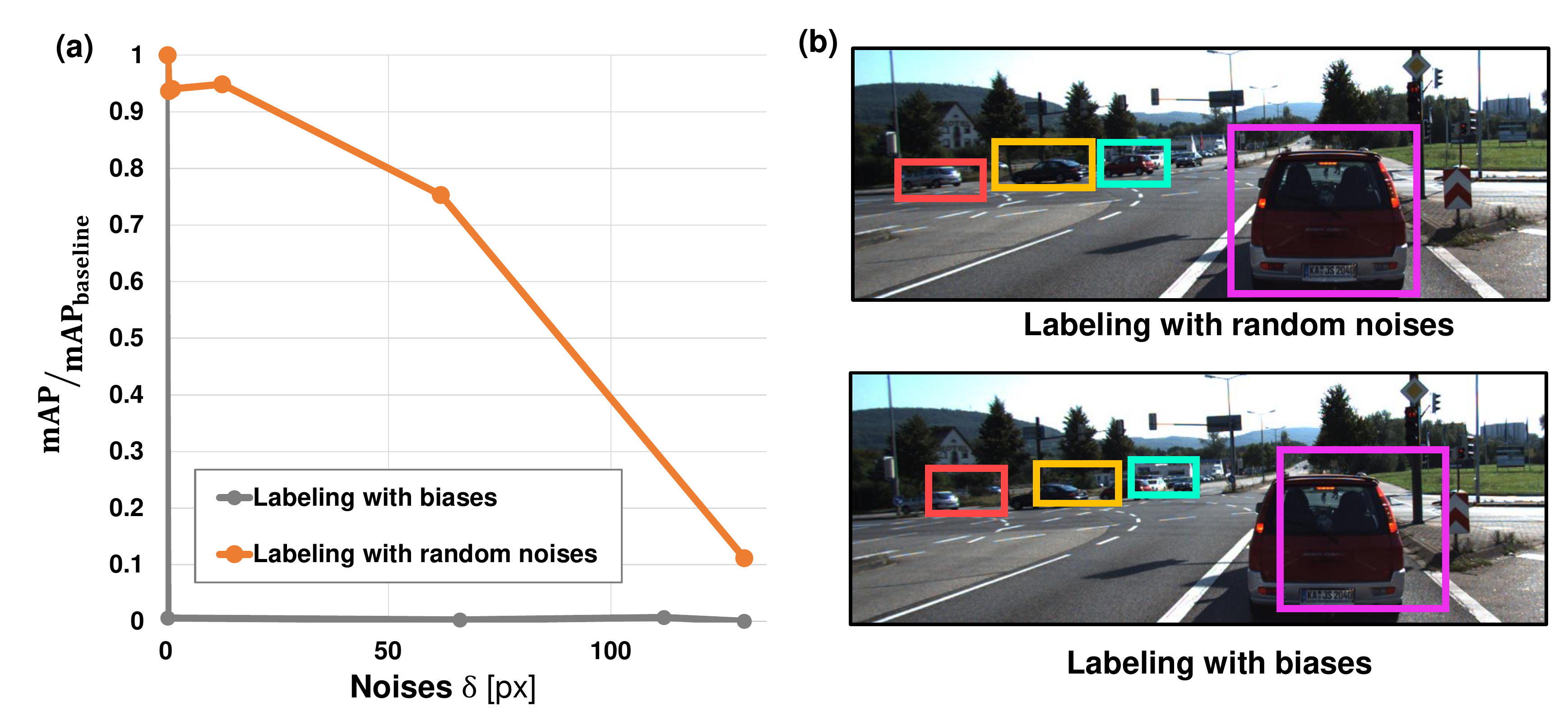}
	\caption{(a) An illustration for the influence of label quality on the performance of an object detection network~\cite{lin2018focal}. The network is trained on labels which are incrementally disturbed. The performance is measured by mAP normalized to the performance trained on the undisturbed dataset. The network is much more robust against random labeling errors (drawn from a Gaussian distribution with variance $\sigma$) than biased labeling (all labels shifted by $\sigma$) cf.~\cite{haase2019estimating, chadwick2019training}. (b) An illustration of the random labeling noises and labeling biases (all bounding boxes are shifted in the upper-right direction).}\label{fig:label_quality}
	\vspace{-0.3cm}
\end{figure}

Besides data diversity and the size of the training dataset, data quality significantly affects the performance of a deep multi-modal perception system as well. Training data is usually labeled by human annotators to ensure the high labeling quality. However, humans are also prone to making errors. Fig.~\ref{fig:label_quality} shows two different errors in the labeling process when training an object detection network. The network is much more robust against labeling errors when they are randomly distributed, compared to biased labeling from the use of a deterministic pre-labeling. Training networks with erroneous labels is further studied in~\cite{wang2018iterative,ren2018learning,jiang2018mentornet,ma2018dimensionality}. The impact on weak or erroneous labels on the performance of deep learning based semantic segmentation is investigated in~\cite{zlateski2018importance, meletis2019boosting}. The influence of labelling errors on the accuracy of object detection is discussed in~\cite{haase2019estimating, chadwick2019training}.

Well-calibrated sensors are the prerequisite for accurate and robust multi-modal perception systems. However, the sensor setup is usually not perfect. Temporal and spatial sensing misalignments might occur while recording the training data or deploying the perception modules. This could cause severe errors in training datasets and degrade the performance of networks, especially for those which are designed to implicitly learn the sensor alignment (e.g. networks that fuse LiDAR BEV feature maps and front view camera images cf. Sec.~\ref{subsubsec:process_lidar_camera}). Interestingly, several works propose to calibrate sensors by deep neural networks: Giering \textit{et al.}~\cite{giering2015multi} discretize the spatial misalignments between LiDAR and visual camera into nine classes, and build a network to classify misalignment taking LiDAR and RGB images as inputs; Schneider \textit{et al.}~\cite{schneider2017regnet} propose to fully regress the extrinsic calibration parameters between LiDAR and visual camera by deep learning. Several multi-modal CNN networks are trained on different de-calibration ranges to iteratively refine the calibration output. In this way, the feature extraction, feature matching, and global optimization problems for sensor registration could be solved in an end-to-end fashion. 

\subsection{Fusion Methodology}\label{subsec:challenge:methodology}
\subsubsection{What to Fuse}\label{subsubsec:challenge:what_to_fuse}
Most reviewed methods combine RGB images with thermal images or LiDAR 3D points. The networks are trained and evaluated on open datasets such as KITTI~\cite{Geiger2012CVPR} and KAIST Pedestrian~\cite{choi2018kaist}. These methods do not specifically focus on sensor redundancy, e.g. installing multiple cameras on a driverless car to increase the reliability of perception systems even when some sensors are defective. How to fuse the sensing information from multiple sensors (e.g. RGB images from multiple cameras) is an important open question. 

Another challenge is how to represent and process different sensing modalities appropriately before feeding them into a fusion network. For instance, many approaches exist to represent LiDAR point clouds, including 3D voxels, 2D BEV maps, spherical maps, as well as sparse or dense depth maps (more details cf. Sec.~\ref{subsec:what_to_fuse}). However, only Pfeuffer \textit{et al.}~\cite{pfeuffer2018optimal} have studied the pros and cons for several LiDAR front-view representations. We expect more works to compare different 3D point representation methods. 

In addition, there are very few studies for fusing LiDAR and camera outputs with signals from other sources such as Radars, ultrasonics or V2X communication. Radar data differs from LiDAR data and it requires different network architecture and fusion schmes. So far, we are not aware of any work fusing Ultrasonic sensor signals in deep multi-modal perception, despite its relevance for low-speed scenarios. How to fuse these sensing modalities and align them temporally and spatially are big challenges. 

Finally, it is an interesting topic to combine physical constraints and model-based approaches with data-driven neural networks. For example, Ramos \textit{et al.}~\cite{ramos2017detecting} propose to fuse semantics and geometric cues in a Bayesian framework for unexpected objects detections. The semantics are predicted by a FCN network, whereas the geometric cues are provided by model-based stereo detections. The multi-task learning scheme also helps to add physical constraints in neural networks. For example, to aid 3D object detection task, Liang \textit{et al.}~\cite{liang2019multi} design a fusion network that additionally estimate LiDAR ground plane and camera image depth. The ground plane estimation provides useful cues for object locations, while the image depth completion contributes to better cross-modal representation; Panoptic segmentation~\cite{kirillov2018panoptic} aims to achieve complete scene understanding by jointly doing semantic segmentation and instance segmentation.

\subsubsection{How to Fuse}\label{subsubsec:challenge:how_to_fuse}
\begin{figure}[tbp]
	\centering
	\centering
	\includegraphics[width=0.98\linewidth]{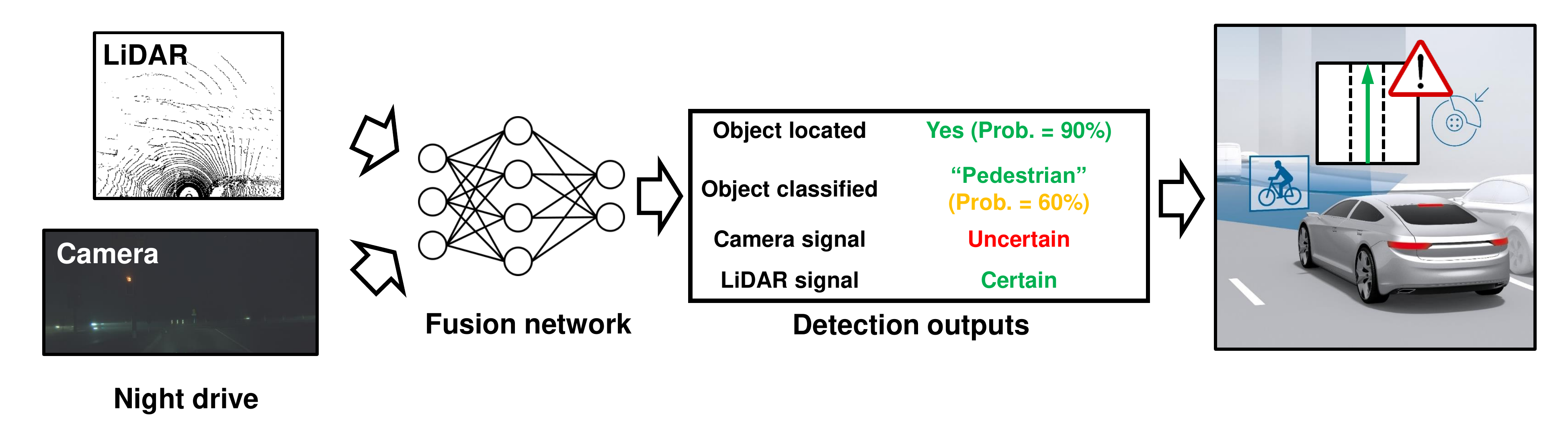}
	\caption{The importance of explicitly modeling and propagating uncertainties in a multi-modal object detection network. Ideally, the network should produce reliable prediction probabilities (object classification and localization). It should e.g. depict high uncertainty for camera signals during a night drive. Such uncertainty information is useful for the decision making modules, such as maneuver planning or emergency braking systems.}\label{fig:uncertainty_propagation}
	\vspace{-0.3cm}
\end{figure}

Explicitly modeling uncertainty or informativeness of each sensing modality is important safe autonomous driving. As an example, a multi-modal perception system should show higher uncertainty against adverse weather or detect unseen driving environments (open-world problem). It should also reflect sensor's degradation or defects as well. The perception uncertainties need to be propagated to other modules such as motion planning~\cite{banzhaf2018footprints} so that the autonomous vehicles can behave accordingly. Reliable uncertainty estimation can show the networks' robustness (cf. Fig~\ref{fig:uncertainty_propagation}).
However, most reviewed papers only fuse multiple sensing modalities by a simple operation (e.g. addition and average mean, cf. Sec.~\ref{subsec:how_to_fuse}). Those methods are designed to achieve high average precision (AP) without considering the networks' robustness. The recent work by Bijelic \textit{et al.}~\cite{bijelic2019seeing} uses dropout to increase the network robustness in foggy images. Specifically, they add pixel-wise dropout masks in different fusion layers so that the network randomly drops LiDAR or camera channels during training. Despite promising results for detections in foggy weather, their method cannot express which sensing modality is more reliable given the distorted sensor inputs. To the best of our knowledge, only the gating network (cf. Sec.~\ref{subsec:how_to_fuse}) explicitly models the informativeness of each sensing modality. 

One way to estimate uncertainty and to increase network robustness is Bayesian Neural Networks (BNNs). They assume a prior distribution over the network weights and infer the posterior distribution to extract the prediction probability~\cite{mackay1992practical}. There are two types of uncertainties BNNs can model. \textit{Epistemic uncertainty} illustrates the models'
uncertainty when describing the training dataset. It can be obtained by estimating the weight posterior by variational inference~\cite{hinton1993keeping}, sampling~\cite{Gal2016Uncertainty,graves2011practical,mandt2017stochastic}, batch normalization~\cite{teye2018bayesian}, or noise injection~\cite{postels2019sampling}. It has been applied to semantic segmentation~\cite{kendall2015bayesian} and open-world object detection problems~\cite{miller2017dropout,miller2018evaluating}. \textit{Aleatoric uncertainty} represents observation noises inherent in sensors. It can be estimated by the observation likelihood such as a Gaussian distribution or Laplacian distribution. Kendall \textit{et al.}~\cite{kendall2017uncertainties} study both uncertainties for semantic segmentation; Ilg \textit{et al.}~\cite{ilg2018uncertainty} propose to extract uncertainties for optical flow; Feng \textit{et al.}~\cite{feng2018towards} examine the epistemic and aleatoric uncertainties in a LiDAR vehicle detection network for autonomous driving. They show that the uncertainties encode very different information. In the successive work, \cite{feng2018leveraging} employ aleatoric uncertainties in a 3D object detection network to significantly improve its detection performance and increase its robustness against noisy data. Other works that introduce aleatoric uncertainties in object detectors include~\cite{meyer2019lasernet,wirges2019capturing,le2018uncertainty,feng2019can}. Although much progress has been made for BNNs, to the best of our knowledge, so far they have not been introduced to multi-modal perception. Furthermore, few works have been done to propagate uncertainties in object detectors and semantic segmentation networks to other modules, such as tracking and motion planning. How to employ these uncertainties to improve the robustness of an autonomous driving system is a challenging open question. 

Another way that can increase the networks' robustness is generative models. In general, generative models aim at modeling the data distribution in an unsupervised way as well as generating new samples with some variations. Variational Autoencoders (VAEs)~\cite{kingma2013auto} and Generative Adversarial Networks (GANs)~\cite{goodfellow2014generative} are the two most popular deep generative models. They have been widely applied to image analysis~\cite{larsen2016autoencoding,deshpande2017learning,isola2017image}, and recently introduced to model Radar data~\cite{wheeler2017deep} and road detection~\cite{han2018semi} for autonomous driving. Generative models could be useful for multi-modal perception problems. For example, they might generate labeled simulated sensor data, when it is tedious and difficult to collect in the real world; they could also serve to detect situations where sensors are defect or an autonomous car is driving into a new scenario that differs from those seen during training. Designing specific fusion operations for deep generative models is an interesting open question. 

\subsubsection{When to Fuse}\label{subsubsec:challenge:when_to_fuse}
As discussed in Sec.~\ref{subsec:when_to_fuse}, the choice of when to fuse the sensing modalities in the reviewed works is mainly based on intuition and empirical results. There is no conclusive evidence that one fusion scheme is better than the others. 
Ideally, the ``optimal'' fusion architecture should be found automatically instead of by meticulous engineering. Neural network structure search can potentially solve the problem. It aims at finding the optimal number of neurons and layers in a neural network. Many approaches have been proposed, including the bottom-up construction approach~\cite{elman1993learning}, pruning~\cite{feng2015learning}, Bayesian optimization~\cite{ramachandram2017structure}, genetic algorithms~\cite{whitley1990genetic}, and the recent reinforcement learning approach~\cite{zoph2016neural}. Another way to optimize the network structure is by regularization, such as $l_1$ regularization~\cite{kulkarni2015learning} and stochastic regularization~\cite{murdock2016blockout,li2017modout}. 

Furthermore, visual analytics techniques could be employed for network architecture design. Such visualization tools can help to understand and analyze how networks behave, to diagnose the problems, and finally to improve the network architecture. Several methods have been proposed for understanding CNNs for image classification~\cite{bilal2018convolutional,liu2017towards}. So far, there has been no research on visual analytics for deep multi-modal learning problems.

\subsubsection{Real-time Consideration}
Deep multi-modal neural networks should perceive driving environments in real-time. Therefore, computational costs and memory requirements should be considered when developing the fusion methodology. At the ``what to fuse'' level, sensing modalities should be represented in an efficient way. At the ``how to fuse'' level, finding fusion operations that are suitable for network acceleration, such as pruning and quantization~\cite{han2015deep,howard2017mobilenets,cheng2017survey,enderich2019learning}, is an interesting future work. At the ``when to fuse'' level, inference time and memory constraints can be considered as regularization term for network architecture optimization.

It is difficult to compare the inference speed among the methods we have reviewed, as there is no benchmark with standard hardware or programming languages. Tab.~\ref{tab:perf_3d} and Tab.~\ref{tab:perf_semseg} summarize the inference speed of several object detection and semantic segmentation networks on the KITTI test set. Each method uses different hardware, and the inference time is reported only by the authors. It is an open question how these methods perform when they are deployed on automotive hardware.


\subsection{Others}
\subsubsection{Evaluation Metrics}
The common way to evaluate object detection methods is mean average precision ($mAP$)~\cite{everingham2010pascal,Geiger2012CVPR}. It is the mean value of average precision (AP) over object classes, given a certain intersection over union ($IoU$) threshold defined as the geometric overlap between predictions and ground truths. As for the pixel-level semantic segmentation, metrics such as average precision, false positive rate, false negative rate, and $IoU$ calculated at pixel level~\cite{cordts2016cityscapes} are often used. However, these metrics only summarize the prediction \textit{accuracy} to a test dataset. They do not consider how sensor behaves in different situations. As an example, to evaluate the performance of a multi-modal network, the $IoU$ thresholds should depend on object distance, occlusion, and types of sensors. 

Furthermore, common evaluation metrics are not designed specifically to illustrate how the algorithm handles open-set conditions or in situations where some sensors are degraded or defective. 
There exist several metrics to evaluate the quality of predictive uncertainty, e.g. empirical calibration curves~\cite{dawid1982well} and log predictive probabilities. The detection error~\cite{liang2017enhancing} measures the effectiveness of a neural network in distinguishing in- and out-of-distribution data. The Probability-based Detection Quality (PDQ)~\cite{hall2018probability} is designed to measure the object detection performance for spatial and semantic uncertainties. These metrics can be adapted to the multi-modal perception problems to compare the networks' robustness.

\subsubsection{More Network Architectures}
Most reviewed methods are based on CNN architectures for single frame perception. The predictions in a frame are not dependent on previous frames, resulting in inconsistency over time. Only a few works incorporate temporal cues (e.g.~\cite{casas2018intentnet,luo2018fast}). Future work is expected to develop multi-modal perception algorithms that can handle time series, e.g. via Recurrent Neural Networks. Furthermore, current methods are designed to propagate results to other modules in autonomous driving, such as localization, planning, and reasoning. While the modular approach is the common pipeline for autonomous driving, some works also try to map the sensor data directly to the decision policy such as steering angles or pedal positions (end-to-end learning)~\cite{bojarski2016end,liu2017learning,bansal2018chauffeurnet}, or to some intermediate environment representations (direct-perception)~\cite{sauer2018conditional,chen2015deepdriving}. Multi-modal end-to-end learning and direct perception can be potential research directions as well. 

\section{\textbf{Conclusion and Discussion}}\label{sec:discussion}
We have presented our survey for deep multi-modal object detection and segmentation applied to autonomous driving. We have provided a summary of both multi-modal datasets and fusion methodologies, considering ``what to fuse'', ``how to fuse'', and ``when to fuse''. We have also discussed challenges and open questions. Furthermore, our interactive online tool allows readers to navigate topics and methods for each reference. We plan to frequently update this tool.   

Despite the fact that an increasing number of multi-modal datasets have been published, most of them record data from RGB cameras, thermal cameras, and LiDARs. Correspondingly, most of the papers we reviewed fuse RGB images either with thermal images or with LiDAR point clouds. Only recently has the fusion of Radar data been investigated. This includes nuScene dataset~\cite{nuscenes2019}, the Oxford Radar RobotCar Dataset~\cite{RadarRobotCarDatasetArXiv}, the Astyx HiRes2019 Dataset~\cite{meyer2019euma}, and the seminal work from Chadwick \textit{et al.}~\cite{chadwick2019distant} that proposes to fuse RGB camera images with Radar points for vehicle detection. In the future, we expect more datasets and fusion methods concerning Radar signals. 

There are various ways to fuse sensing modalities in neural networks, encompassing different sensor representations, cf. Sec.~\ref{subsec:what_to_fuse}, fusion operations cf. Sec.~\ref{subsec:how_to_fuse}, and fusion stages, cf. Sec.~\ref{subsec:when_to_fuse}. However, we do not find conclusive evidence that one fusion method is better than the others. Additionally, there is a lack of research on multi-modal perception in open-set conditions or with sensor failures. We expect more focus on these challenging research topics. 
\section*{Acknowledgment}
We thank Fabian Duffhauss for collecting literature and reviewing the paper. We also thank Bill Beluch, Rainer Stal, Peter M\"oller and Ulrich Michael for their suggestions and inspiring discussions. 
\bibliographystyle{IEEEtran}

\bibliography{IEEEabrv,bibliography}
\begin{IEEEbiography}[{\includegraphics[width=1in,height=1.25in,clip,keepaspectratio]{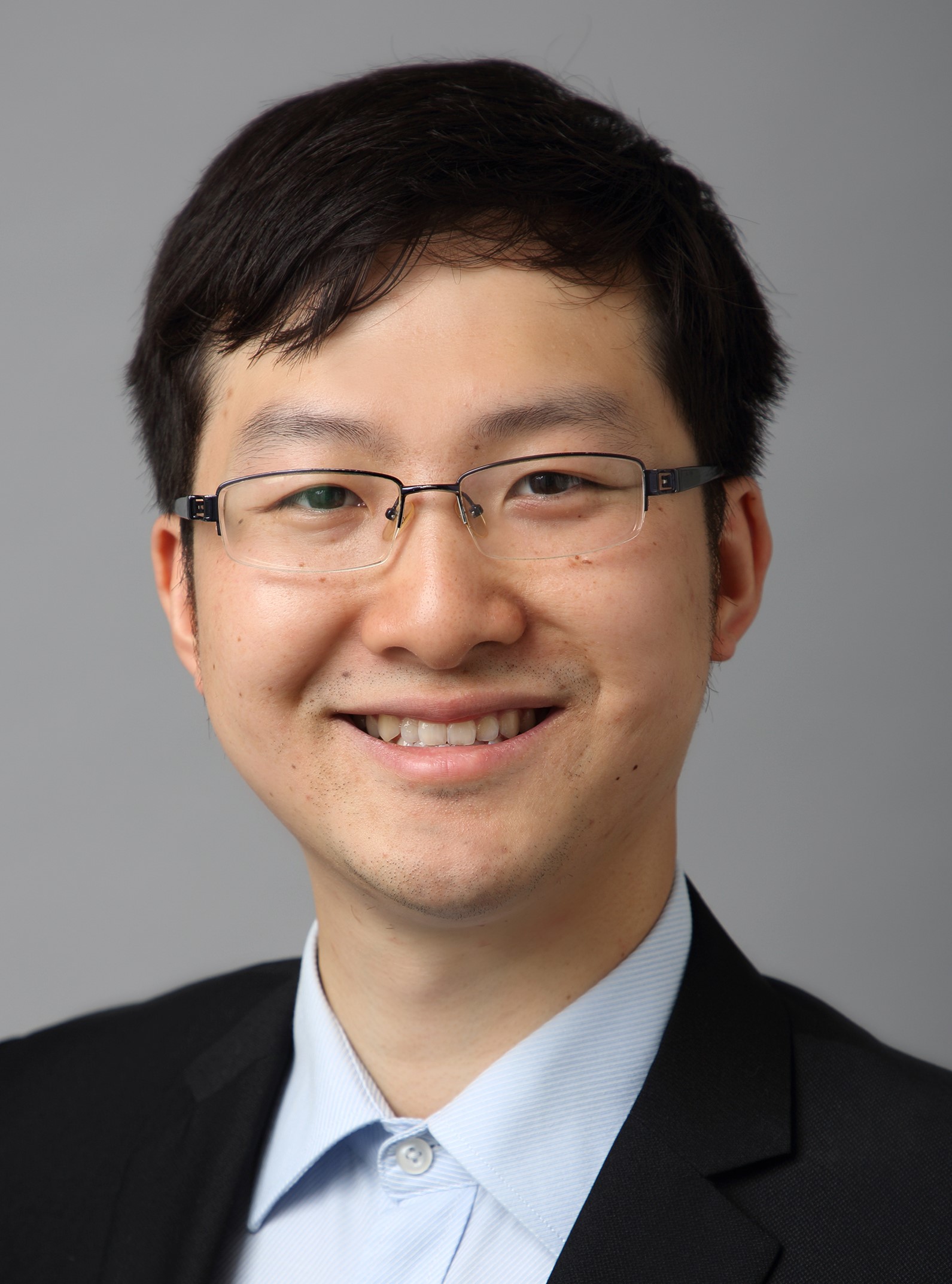}}]{Di Feng}
(Member, IEEE) is currently pursuing his doctoral degree in the Corporate Research of Robert Bosch GmbH, Renningen, in cooperation with the Ulm University. He finished his master's degree with distinction in electrical and computer engineering at the Technical University of Munich. During his studies, he was granted the opportunity to work in several teams with reputable companies and research institutes such as BMW AG, German Aerospace Center (DLR), and Institute for Cognitive Systems (ICS) at Technical University of Munich. He received the bachelor's degree in mechatronics with honor from Tongji University. His current research is centered on robust multi-modal object detection using deep learning approach for autonomous driving. He is also interested in robotic active learning and exploration through tactile sensing and cognitive systems. 
\end{IEEEbiography}
\vskip -1cm
\begin{IEEEbiography}
[{\includegraphics[width=1in,height=1.25in,clip,keepaspectratio]{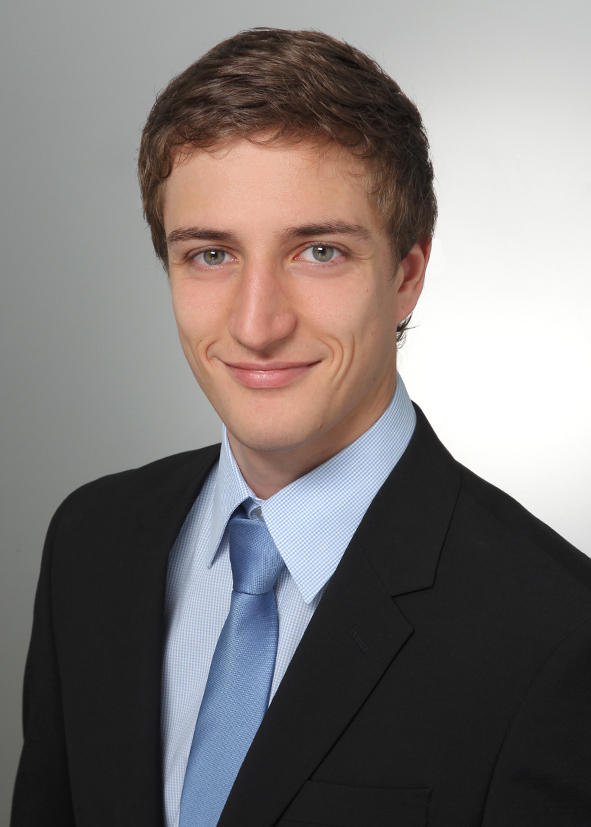}}]{Christian Haase-Sch\"utz} (Member, IEEE) is currently pursuing his PhD degree at Chassis Systems Control, Robert Bosch GmbH, Abstatt, in cooperation with the Karlsruhe Institute of Technology. Before joining Bosch, he finished his master's degree in physics at the Friedrich-Alexander-University Erlangen-Nuremberg. He did his thesis with the Center for Medical Physics. During his master studies he was granted a scholarship by the Bavarian state to visit Huazhong University of Science and Technology, Wuhan, China, from March 2015 till July 2015. He received his bachelor's degree in physics from the University of Stuttgart in 2013, where he did his thesis with the Max Planck Institute for Intelligent Systems. His professional experience includes work with ETAS GmbH, Stuttgart, and Andreas Stihl AG, Waiblingen. His current research is centered on multi-modal object detection using deep learning approaches for autonomous driving. He is further interested in challenges of AI systems in the wild. Christian Haase-Sch\"utz is a member of the IEEE and the German Physical Society DPG.
\end{IEEEbiography}
\vskip -1cm
\begin{IEEEbiography}
[{\includegraphics[width=1in,height=1.25in,clip,keepaspectratio]{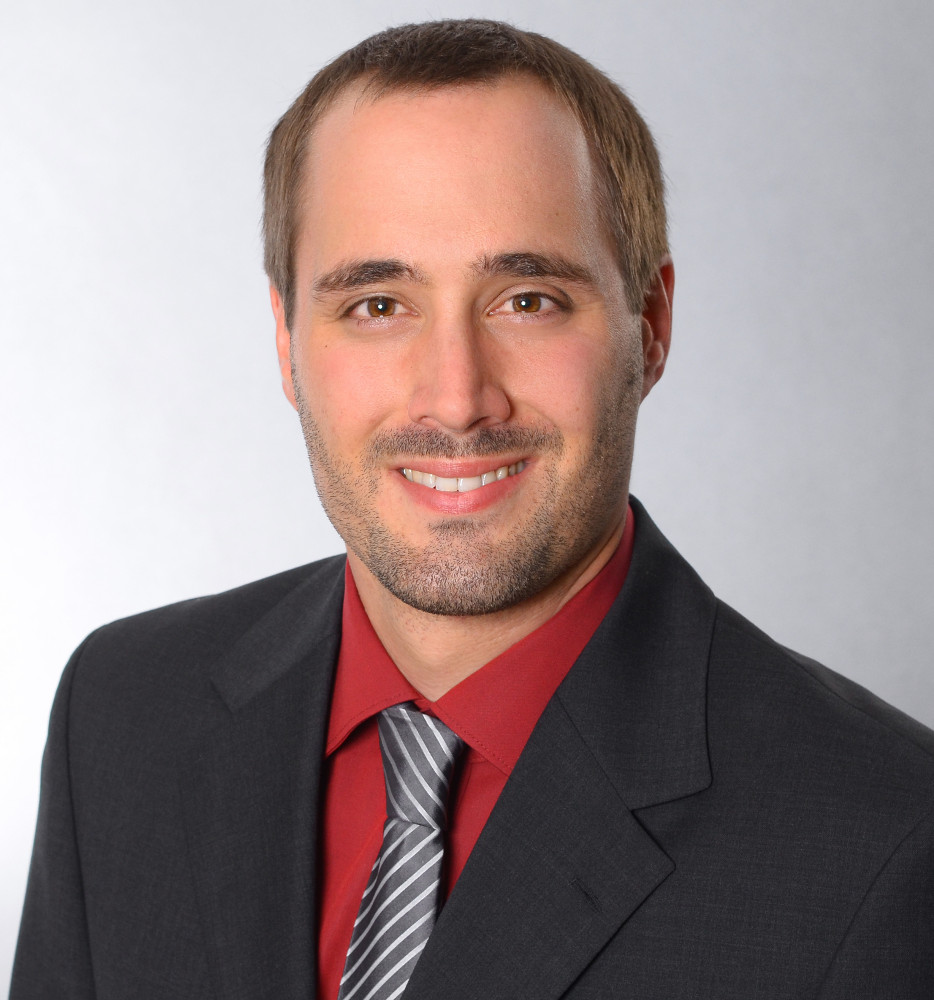}}]{Lars Rosenbaum} received his Dipl.-Inf. (M.S.) and the Dr. rer. nat. (Ph.D.) degrees in bioinformatics from the University of Tuebingen, Germany, in 2009 and 2013, respectively. During this time he was working on machine learning approaches for computer-aided molecular drug design and analysis of metabolomics data. In 2014, he joined ITK Engineering in Marburg, Germany, working on driver assistance systems. Since 2016, he is a research engineer at Corporate Research, Robert Bosch GmbH in Renningen, Germany, where he is currently doing research on machine learning algorithms in the area of perception for automated driving.
\end{IEEEbiography}
\vskip -1cm
\begin{IEEEbiography}
[{\includegraphics[width=0.8in,height=1.25in,clip,keepaspectratio]{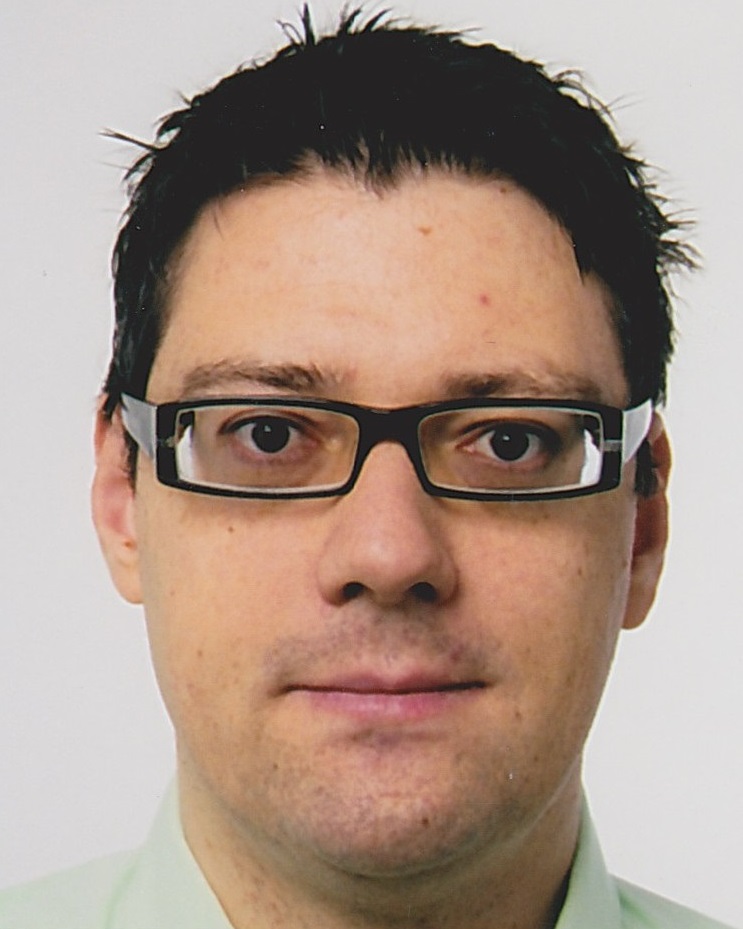}}]{Heinz Hertlein} (Member, IEEE) received the Dipl.-Inf. degree (diploma in computer science) from the Friedrich-Alexander-University Erlangen-Nuremberg, Germany, in 1999, and the Dr.-Ing. (Ph.D.) degree from the same university in 2010 for his research in the field of biometric speaker recognition. From 2002, he was working on algorithms and applications of multi modal biometric pattern recognition at the company BioID in Erlangen-Tennenlohe and Nuremberg. From 2012, he was appointed at the University of Hertfordshire in the UK, initially as a Postdoctoral Research Fellow and later as a Senior Lecturer. He was teaching in the fields of signal processing and pattern recognition, and his research activities were mainly focused on biometric speaker and face recognition. Since 2015, he is employed at Chassis Systems Control, Robert Bosch GmbH in Abstatt, Germany, where he is currently working in the area of perception for autonomous driving.
\end{IEEEbiography}
\vskip -1cm
\begin{IEEEbiography}[{\includegraphics[width=1in,height=1.25in,clip,keepaspectratio]{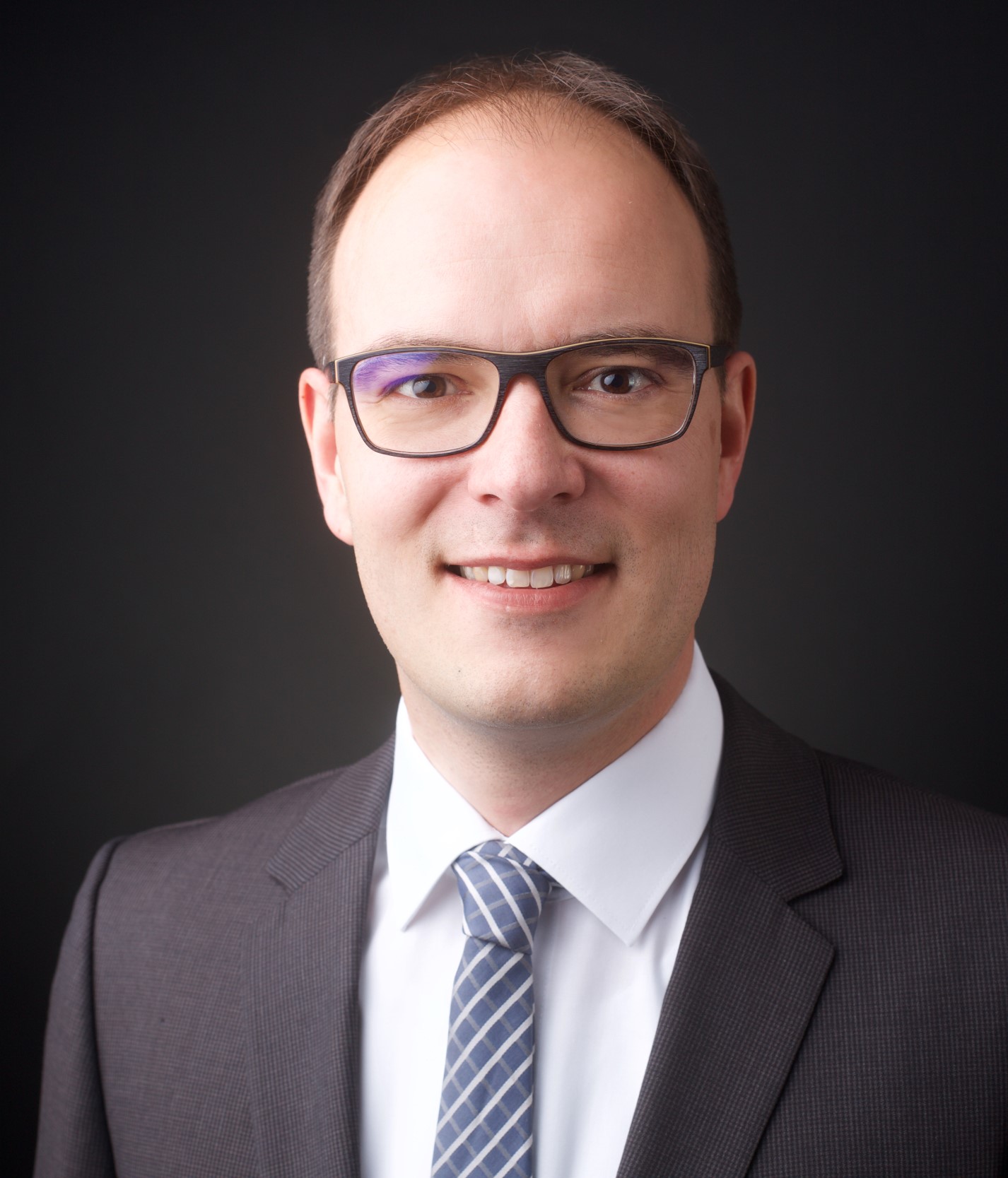}}]{Claudius Gl\"aser} was born in Gera, Germany in 1982. He received his Diploma degree in Computer Science from the Technical University of Ilmenau, Germany, in 2006, and the Dr.-Ing. degree (equivalent to PhD) from Bielefeld University, Germany, in 2012. From 2006 he was a Research Scientist at the Honda Research Institute Europe GmbH in Offenbach/Main, Germany, working in the fields of speech processing and language understanding for humanoid robots. In 2011 he joined the Corporate Research of Robert Bosch GmbH in Renningen, Germany, where he developed perception algorithms for driver assistance and highly automated driving functions. Currently, he is Team Lead for perception for automated driving and manages various related projects. His research interests include environment perception, multimodal sensor data fusion, multi-object tracking, and machine learning for highly automated driving.
\end{IEEEbiography}
\vskip -1cm
\begin{IEEEbiography}[{\includegraphics[width=1in,height=1.25in,clip,keepaspectratio]{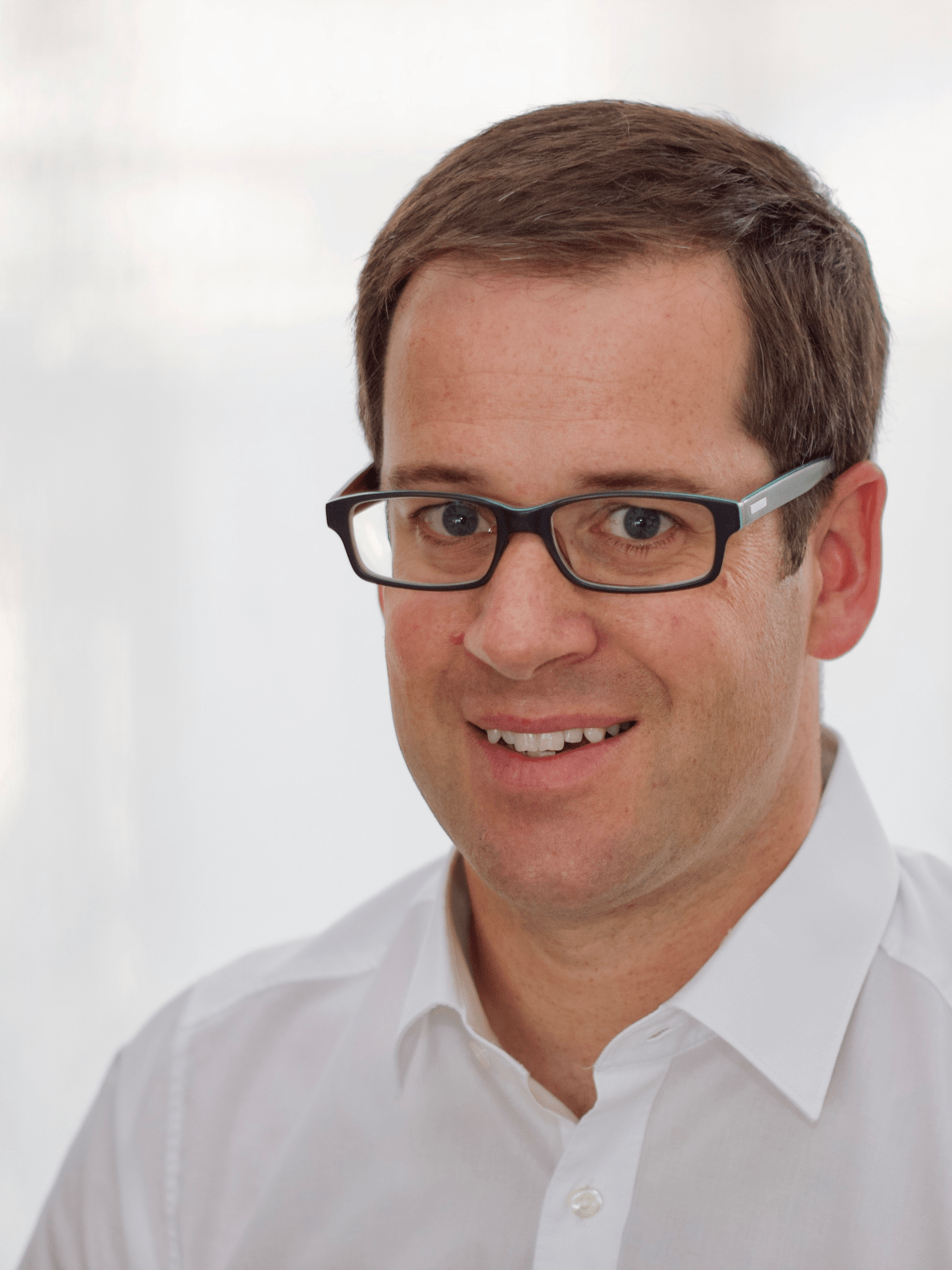}}]{Fabian Timm} studied Computer Science at the University of L\"ubeck, Germany. In 2006 he did his diploma thesis at Philips Lighting Systems in Aachen, Germany. Afterwards he started his PhD also at Philips Lighting Systems in the field of Machine Vision and Machine Learning and finished it in 2011 at the University of L\"ubeck, Institute for Neuro- and Bioinformatics. In 2012 he joined corporate research at Robert Bosch GmbH and worked on industrial image processing and machine learning. Afterwards he worked in the business unit at Bosch and developed new perception algorithms, such as pedestrian and cyclist protection only with a single radar sensor. Since 2018 he leads the research group "automated driving – perception and sensors" at Bosch corporate research. His main research interests are machine and deep learning, signal processing, and sensors for automated driving.
\end{IEEEbiography}
\vskip -1cm
\begin{IEEEbiography}
[{\includegraphics[width=1in,height=1.25in,clip,keepaspectratio]{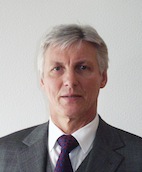}}]{Werner Wiesbeck} (Fellow, IEEE) 
received the Dipl.-Ing. (M.S.) and the Dr. -Ing. (Ph.D.) degrees in electrical engineering from the Technical University Munich, Germany, in 1969 and 1972, respectively. From 1972 to 1983, he was with product responsibility for mm-wave radars, receivers, direction finders, and electronic    warfare systems in industry. From 1983 to 2007, he was the Director of the Institut f\"ur H\"ochstfrequenztechnik und  Elektronik, University of Karlsruhe. He is currently a Distinguished Senior Fellow at the Karlsruhe Institute of Technology. His research topics include antennas, wave propagation, radar, remote sensing, wireless communication, and ultra wide band technologies.\\
He has authored and co-authored several books and over 850 publications, is a supervisor of over 90 Ph.D. students, a responsible supervisor of over 600 Diploma-/Master theses, and holds over 60 patents. He is an Honorary Life Member of IEEE GRS-S and a member of the Heidelberger Academy of Sciences and Humanities, and the German Academy of Engineering and Technology. He was a recipient of a number of awards, including the IEEE Millennium Award, the IEEE GRS Distinguished Achievement Award, the Honorary Doctorate (Dr. h. c.) from the University Budapest/Hungary, the Honorary Doctorate (Dr.-Ing. E. h.) from the University Duisburg/Germany, the Honorary Doctorate (Dr. -Ing. E. h.) from Technische Universit\"at Ilmenau, and the IEEE Electromagnetics Award in 2008. He is the Chairman of the GRS-S Awards Committee. He was the Executive Vice President of IEEE GRS-S (1998-1999) and the President of IEEE GRS-S (2000-2001). He has been a general chairman of several conferences.

\end{IEEEbiography}
\vskip -1cm
\begin{IEEEbiography}
[{\includegraphics[width=1in,height=1.25in,clip,keepaspectratio]{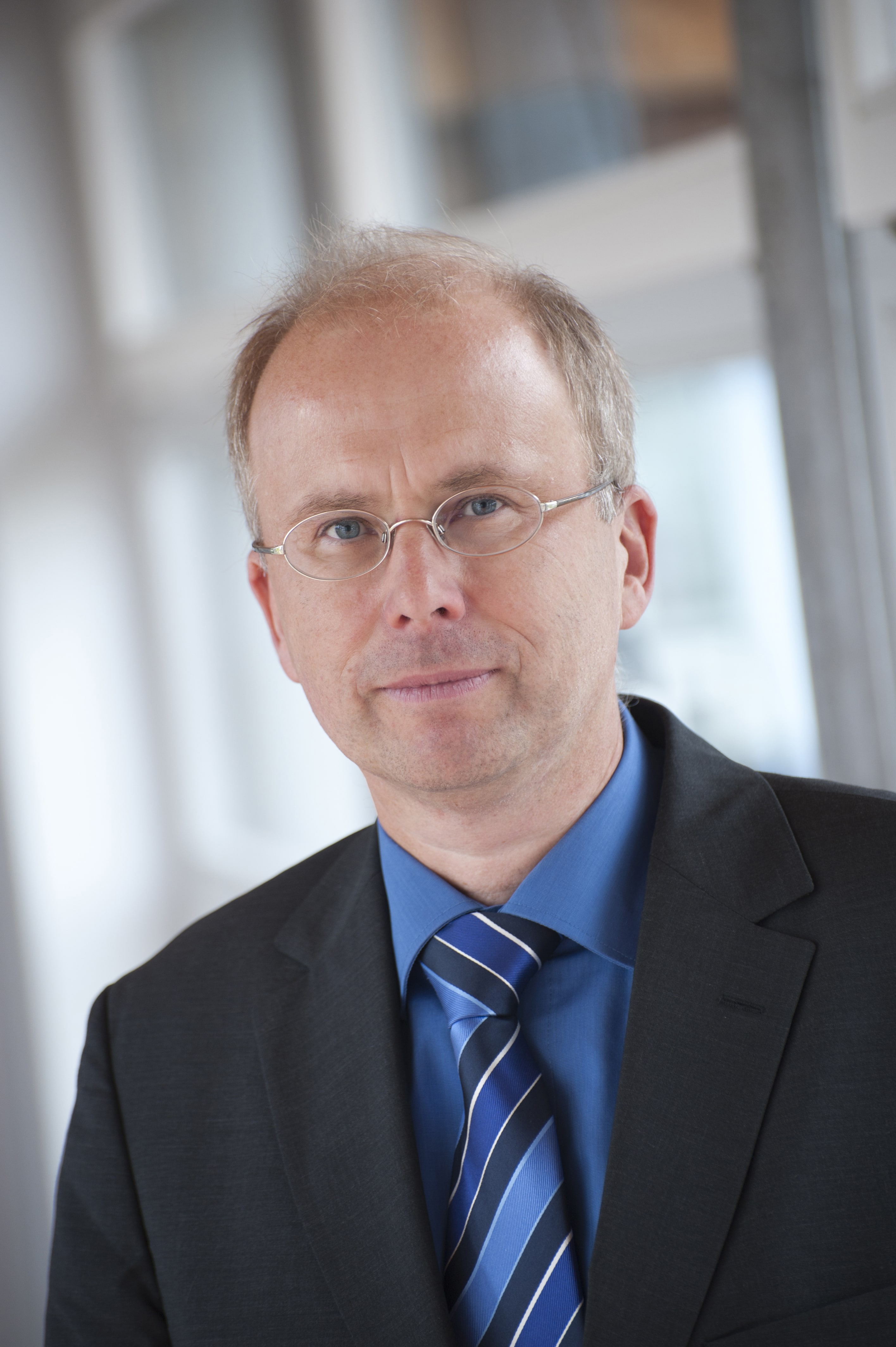}}]{Klaus Dietmayer} (Member, IEEE) was born in Celle, Germany in 1962. He received his Diploma degree in 1989 in Electrical Engineering from the Technical University of Braunschweig (Germany), and the Dr.-Ing. degree (equivalent to PhD) in 1994 from the University of Armed Forces in Hamburg (Germany). In 1994 he joined the Philips Semiconductors Systems Laboratory in Hamburg, Germany as a research engineer. Since 1996 he became a manager in the field of networks and sensors for automotive applications. In 2000 he was appointed to a professorship at the University of Ulm in the field of measurement and control. Currently he is Full Professor and Director of the Institute of Measurement, Control and Microtechnology in the school of Engineering and Computer Science at the University of Ulm. Research interests include information fusion, multi-object tracking, environment perception, situation understanding and trajectory planning for autonomous driving. Klaus Dietmayer is member of the IEEE and the German society of engineers VDI / VDE.
\end{IEEEbiography}

\onecolumn
\begin{landscape}
\captionsetup{font=normalsize}
\scriptsize
\centering
\begin{ThreePartTable}


\begin{tabularx}{1.0\linewidth}{>{\raggedright\arraybackslash\hsize=0.5\hsize}X >{\raggedright\arraybackslash\hsize=0.4\hsize}X >{\raggedright\arraybackslash\hsize=0.2\hsize}X >{\raggedright\arraybackslash\hsize=0.4\hsize}X >{\raggedright\arraybackslash\hsize=0.3\hsize}X >{\raggedright\arraybackslash\hsize=0.5\hsize}X >{\raggedright\arraybackslash\hsize=0.6\hsize}X >{\raggedright\arraybackslash\hsize=0.7\hsize}X >{\hsize=0.0\hsize}X}
\caption{OVERVIEW OF MULTI-MODAL DATASETS}\label{tab:datasets}\\
\rowcolor{lightgray!75}
\textbf{Name} & \textbf{Sensing Modalities} & \textbf{Year (published)} & \textbf{Labelled (benchmark)} & \textbf{Recording area} & \textbf{Size} & \textbf{Categories / Remarks} & \textbf{Link} \\[6pt]

\endhead
\endfoot

Astyx HiRes2019 \cite{meyer2019euma} & Radar, Visual camera, 3D LiDAR & 2019 & 3D bounding boxes & n.a. & 500 frames (5000 annotated objects) & Car, Bus, Cyclist, Motorcyclist, Person, Trailer, Truck & \url{https://www.astyx.com/development/astyx-hires2019-dataset.html}\\ \midrule

A2D2 \cite{aev2019} & Visual cameras (6); 3D LiDAR (5); Bus data & 2019 & 2D/3D bounding boxes, 2D/3D instance segmentation & Gaimersheim, Ingolstadt, Munich & 40k frames (semantics), 12k frames (3D objects), 390k frames unlabeled & Car, Bicycle, Pedestrian, Truck, Small vehicles, Traffic signal, Utility vehicle, Sidebars, Speed bumper, Curbstone, Solid line, Irrelevant signs, Road blocks, Tractor, Non-drivable street, Zebra crossing, Obstacles / trash, Poles, RD restricted area, Animals, Grid structure, Signal corpus, Drivable cobbleston, Electronic traffic, Slow drive area, Nature object, Parking area, Sidewalk, Ego car, Painted driv. instr., Traffic guide obj., Dashed line, RD normal street, Sky, Buildings, Blurred area, Rain dirt & \url{https://www.audi-electronics-venture.de/aev/web/en/driving-dataset.html}\\ \midrule

A*3D Dataset \cite{astar-3d} & Visual cameras (2); 3D LiDAR & 2019 & 3D bounding boxes & Singapore & 39k frames, 230k objects & Car, Van, Bus, Truck, Pedestrians, Cyclists, and Motorcyclists; Afternoon and night, wet and dry & \url{https://github.com/I2RDL2/ASTAR-3D}\\ \midrule

EuroCity Persons \cite{braun2019eurocity} & Visual camera; Announced: stereo, LiDAR, GNSS and intertial sensors & 2019 & 2D bounding boxes & 12 countries in Europe, 27 cities & 47k frames, 258k objects & Pedestrian, Rider, Bicycle, Motorbike, Scooter, Tricycle, Wheelchair, Buggy, Co-Rider; Highly diverse: 4 seasons, day and night, wet and dry & \url{https://eurocity-dataset.tudelft.nl/eval/overview/home}\\ \midrule

Oxford RobotCar \cite{RobotCarDatasetIJRR, RadarRobotCarDatasetArXiv} & 2016: Visual cameras (fisheye \& stereo), 2D \& 3D LiDAR, GNSS, and inertial sensors; 2019: Radar, 3D Lidar (2), 2D LiDAR (2), visual cameras (6), GNSS and inertial sensors & 2016, 2019 & no & Oxford & 2016: $\unit[11,070,651]{\ frames}$ (stereo), $\unit[3,226,183]{frames}$ (3D LiDAR); 2019: 240k scans (Radar), 2.4M frames (LiDAR) & Long-term autonomous driving. Various weather conditions, including heavy rain, night, direct sunlight and snow. & \url{http://robotcar-dataset.robots.ox.ac.uk/downloads/}, \url{http://ori.ox.ac.uk/datasets/radar-robotcar-dataset} \\  \midrule

Waymo Open Dataset \cite{waymo_open_dataset} & 3D LiDAR (5), Visual cameras (5) & 2019 & 3D bounding box, Tracking & n.a. & 200k frames, 12M objects (3D LiDAR), 1.2M objects (2D camera) & Vehicles, Pedestrians, Cyclists, Signs & \url{https://waymo.com/open/}\\	\midrule

Lyft Level 5 AV Dataset 2019 \cite{lyft2019} & 3D LiDAR (5), Visual cameras (6) & 2019 & 3D bounding box & n.a. & 55k frames & Semantic HD map included & \url{https://level5.lyft.com/dataset/}\\	\midrule

Argoverse \cite{Chang_2019_CVPR} & 3D LiDAR (2), Visual cameras (9, 2 stereo) & 2019 & 3D bounding box, Tracking, Forecasting & Pittsburgh, Pennsylvania, Miami, Florida & 113 scenes, 300k trajectories & Vehicle, Pedestrian, Other Static, Large Vehicle, Bicycle, Bicyclist, Bus, Other Mover, Trailer, Motorcyclist, Moped, Motorcycle, Stroller, Emergency Vehicle, Animal, Wheelchair, School Bus; Semantic HD maps (2) included & \url{https://www.argoverse.org/data.html}\\	\midrule

PandaSet \cite{pandaset} & 3D LiDAR (2), Visual cameras (6), GNSS and inertial sensors & 2019 & 3D bounding box & San Francisco, El Camino Real & Announced: 60k frames (camera), 20k frames (LiDAR), 125 scenes & 28 classes, 37 semantic segmentation labels; Solid state LiDAR & \url{https://scale.com/open-datasets/pandaset}\\ \midrule

nuScenes dataset \cite{nuscenes2019} & Visual cameras (6), 3D LiDAR, and Radars (5) & 2019
& 3D bounding box & Boston, Singapore & 1000 scenes, 1.4M frames (camera, Radar), 390k frames (3D LiDAR) & 25 Object classes, such as Car / Van / SUV, different Trucks, Buses, Persons, Animal, Traffic Cone, Temporary Traffic Barrier, Debris, etc. & \url{https://www.nuscenes.org/download} \\  \midrule

BLVD \cite{blvdICRA2019} & Visual (Stereo) camera, 3D LiDAR & 2019
& 3D bounding box, Tracking, Interaction, Intention & Changshu & 120k frames,  $\unit[249,129]{\ objects}$ & Vehicle, Pedestrian, Rider during day and night & \url{https://github.com/VCCIV/BLVD/} \\  \midrule

H3D dataset \cite{360LiDARTracking_ICRA_2019} & Visual cameras (3), 3D LiDAR & 2019 & 3D bounding box & San Francisco, Mountain View, Santa Cruz, San Mateo & $\unit[27,721]{\ frames}$, $\unit[1,071,302]{\ objects}$ & Car, Pedestrian, Cyclist, Truck, Misc, Animals, Motorcyclist, Bus & \url{https://usa.honda-ri.com/hdd/introduction/h3d} \\  \midrule


ApolloScape \cite{apolloscape_arXiv_2018} & Visual (Stereo) camera, 3D LiDAR, GNSS, and inertial sensors & 2018, 2019
& 2D/3D pixel-level segmentation, lane marking, instance segmentation, depth & Multiple areas in China & $\unit[143,906]{\ image\ frames}$, $\unit[89,430]{\ objects}$& Rover, Sky, Car, Motobicycle, Bicycle, Person, Rider, Truck, Bus, Tricycle, Road, Sidewalk, Traffic Cone, Road Pile, Fence, Traffic Light, Pole, Traffic Sign, Wall, Dustbin, Billboard, Building, Bridge, Tunnel, Overpass, Vegetation & \url{http://apolloscape.auto/scene.html} \\  \midrule

DBNet Dataset \cite{chen2018lidar} & 3D LiDAR, Dashboard visual camera, GNSS & 2018 & Driving behaviours (Vehicle speed and wheel angles) & Multiple areas in China & Over 10k frames & In total seven datasets with different test scenarios, such as seaside roads, school areas, mountain roads. & \url{http://www.dbehavior.net/} \\  \midrule

KAIST multispectral dataset \cite{choi2018kaist} & Visual (Stereo) and thermal camera, 3D LiDAR, GNSS, and inertial sensors & 2018 & 2D bounding box, drivable region, image enhancement, depth, and colorization & Seoul & $\unit[7,512]{\ frames}$, $\unit[308,913]{\ objects}$ & Person, Cyclist, Car during day and night, fine time slots (sunrise, afternoon,...) & \url{http://multispectral.kaist.ac.kr} \\  \midrule

Multi-spectral Object Detection dataset \cite{takumi2017multispectral} & Visual and thermal cameras & 2017 & 2D bounding box & University environment in Japan & $\unit[7,512]{\ frames}$, $\unit[5,833]{\ objects}$ & Bike, Car, Car Stop, Color Cone, Person during day and night& \url{https://www.mi.t.u-tokyo.ac.jp/static/projects/mil_multispectral/} \\  \midrule

Multi-spectral Semantic Segmentation dataset \cite{ha2017mfnet} & Visual and thermal camera & 2017 & 2D pixel-level segmentation & n.a. & $\unit[1569]{\ frames}$ & Bike, Car, Person, Curve, Guardrail, Color Cone, Bump during day and night & \url{https://www.mi.t.u-tokyo.ac.jp/static/projects/mil_multispectral/} \\  \midrule

Multi-modal Panoramic 3D Outdoor (MPO) dataset \cite{jung2016multi} & Visual camera, LiDAR, and GNSS & 2016 & Place categorization & Fukuoka & $\unit[650]{\ scans}$ (dense), $\unit[34200]{\ scans}$ (sparse) & No dynamic objects & \url{http://robotics.ait.kyushu-u.ac.jp/~kurazume/research-e.php?content=db\#d08} \\  \midrule

KAIST multispectral pedestrian \cite{hwang2015multispectral}  & Visual and thermal camera & 2015 & 2D bounding box & Seoul & $\unit[95,328]{\ frames}$, $\unit[103,128]{\ objects}$ & Person, People, Cyclist during day and night& \url{https://sites.google.com/site/pedestrianbenchmark/home} \\ \midrule

KITTI \cite{Geiger2012CVPR, Geiger2013IJRR} & Visual (Stereo) camera, 3D LiDAR, GNSS, and inertial sensors & 2012, 2013, 2015 & 2D/3D bounding box, visual odometry, road, optical flow, tracking, depth, 2D instance and pixel-level segmentation & Karlsruhe & $\unit[7481]{\ frames}$ (training) $\unit[80.256]{\ objects}$ & Car, Van, Truck, Pedestrian, Person (sitting), Cyclist, Tram, Misc & \url{http://www.cvlibs.net/datasets/kitti/} \\  \midrule

The M\'alaga Stereo and Laser Urban dataset \cite{blanco2014malaga} & Visual (Stereo) camera, 5$\times$ 2D LiDAR (yielding 3D information), GNSS and inertial sensors & 2014 & no & M\'alaga & $\unit[113,082]{frames}$, $\unit[5,654.6]{s}$  (camera); $>\unit[220,000]{frames}$, $\unit[~5,000]{s}$  (LiDARs) & n.a. & \url{https://www.mrpt.org/MalagaUrbanDataset} \\  
\bottomrule

\end{tabularx}

\end{ThreePartTable}
\captionsetup{font=normalsize}
\scriptsize
\centering
\begin{ThreePartTable}

\begin{tabularx}{1.00\linewidth}{>{\raggedright\arraybackslash\hsize=0.3\hsize}X >{\raggedright\arraybackslash\hsize=0.5\hsize}X >{\raggedright\arraybackslash\hsize=0.5\hsize}X >
{\raggedright\arraybackslash\hsize=0.9\hsize}X >{\raggedright\arraybackslash\hsize=0.2\hsize}X >{\raggedright\arraybackslash\hsize=0.7\hsize}X >{\raggedright\arraybackslash\hsize=0.3\hsize}X >{\raggedright\arraybackslash\hsize=0.7\hsize}X >{\raggedright\arraybackslash\hsize=0.2\hsize}X >
{\raggedright\arraybackslash\hsize=0.7\hsize}X >{\hsize=0.0\hsize}X}
\caption{SUMMARY OF MULTI-MODAL OBJECT DECTECTION METHODS}\label{tab:object_detection_summary}\\
\rowcolor{lightgray!75}
\textbf{Reference} & \textbf{Sensors} & \textbf{Obj Type} & \textbf{Sensing Modality Representations and Processing} & \textbf{Network Pipeline}  & \textbf{How to generate Region Proposals (RP) \tnote{a}} & \textbf{When to fuse} & \textbf{Fusion Operation and Method} & \textbf{Fusion Level\tnote{b}} & \textbf{Dataset(s) used} \\[4pt]

\endhead
\endfoot

Liang \textit{et al.}, 2019\cite{liang2019multi} & LiDAR, visual camera & 3D Car, Pedestrian, Cyclist  & LiDAR BEV maps, RGB image. Each processed by a ResNet with auxiliary tasks: depth estimation and ground segmentation & Faster R-CNN & Predictions with fused features & Before RP & Addition, continuous fusion layer & Middle & KITTI, self-recorded \\ \midrule

Wang \textit{et al.}, 2019\cite{wang2019frustum} & LiDAR, visual camera & 3D Car, Pedestrian, Cyclist, Indoor objects & LiDAR voxelized frustum (each frustum processed by the PointNet), RGB image (using a pre-trained detector). & R-CNN & Pre-trained RGB image detector & After RP & Using RP from RGB image detector to build LiDAR frustums & Late & KITTI, SUN-RGBD \\ \midrule

Dou \textit{et al.}, 2019\cite{dou2019seg} & LiDAR, visual camera & 3D Car & LiDAR voxel (processed by VoxelNet), RGB image (processed by a FCN to get semantic features) & Two stage detector & Predictions with fused features & Before RP & Feature concatenation & Middle & KITTI  \\ \midrule

Sindagi \textit{et al.}, 2019\cite{sindagi2019mvx} & LiDAR, visual camera & 3D Car & LiDAR voxel (processed by VoxelNet), RGB image (processed by a pre-trained 2D image detector). & One stage detector & Predictions with fused features & Before RP & Feature concatenation & \textbf{Early}, Middle & KITTI  \\ \midrule

Bijelic \textit{et al.}, 2019\cite{bijelic2019seeing} & LiDAR, visual camera & 2D Car in foggy weather & Lidar front view images (depth, intensity, height), RGB image. Each processed by VGG16 & SSD & Predictions with fused features & Before RP & Feature concatenation & From early to middle layers & Self-recorded datasets focused on foggy weather, simulated foggy images from KITTI  \\ \midrule

Chadwick \textit{et al.}, 2019\cite{chadwick2019distant} & Radar, visual camera & 2D Vehicle & Radar range and velocity maps, RGB image. Each processed by ResNet & One stage detector & Predictions with fused features & Before RP & Addition, feature concatenation & Middle & Self-recorded \\ \midrule

Liang \textit{et al.}, 2018\cite{liang2018deep} & LiDAR, visual camera & 3D Car, Pedestrian, Cyclist & LiDAR BEV maps, RGB image. Each processed by ResNet & One stage detector & Predictions with fused features & Before RP & Addition, continuous fusion layer & Middle & KITTI, self-recorded \\ \midrule

Du \textit{et al.}, 2018\cite{du2018general} & LiDAR, visual camera & 3D Car & LiDAR voxel (processed by RANSAC and model fitting), RGB image (processed by VGG16 and GoogLeNet) & R-CNN & Pre-trained RGB image detector produces 2D bounding boxes to crop LiDAR points, which are then clustered & Before and at RP & Ensemble: use RGB image detector to regress car dimensions for a model fitting algorithm & Late & KITTI, self-recorded data \\ \midrule

Kim \textit{et al}, 2018\cite{kim2018robust} & LiDAR, visual camera & 2D Car & LiDAR front-view depth image, RGB image Each input processed by VGG16 & SSD & SSD with fused features & Before RP & Feature concatenation, Mixture of Experts & Middle & KITTI \\ \midrule

Yang \textit{et al.}, 2018\cite{yang2018hdnet} & LiDAR, HD-map & 3D Car & LiDAR BEV maps, Road mask image from HD map. Inputs processed by PIXOR++~\cite{yang2018pixor} with the backbone similar to FPN & One stage detector & Detector predictions & Before RP & Feature concatenation & Early & KITTI, TOR4D Dataset~\cite{yang2018pixor}\\ \midrule

Pfeuffer \textit{et al.}, 2018\cite{pfeuffer2018optimal} & LiDAR, visual camera & Multiple 2D objects & LiDAR spherical, and front-view sparse depth, dense depth image, RGB image. Each processed by VGG16 & Faster R-CNN & RPN from fused features & Before RP & Feature concatenation & Early, \textbf{Middle}, Late & KITTI \\ \midrule

Casas \textit{et al.}, 2018\cite{casas2018intentnet}\tnote{c} & LiDAR, HD-map & 3D Car & sequential LiDAR BEV maps, sequential several road topology mask images from HD map. Each input processed by a base network with residual blocks & One stage detector & Detector predictions & Before RP & Feature concatenation & Middle & self-recorded data \\ \midrule

Guan \textit{et al.}, 2018\cite{guan2018fusion} & visual camera, thermal camera & 2D Pedestrian & RGB image, thermal image. Each processed by a base network built on VGG16 & Faster R-CNN & RPN with fused features & Before and after RP & Feature concatenation, Mixture of Experts & Early,  Middle, Late & KAIST Pedestrian Dataset \\ \midrule

Shin \textit{et al.}, 2018\cite{shin2018roarnet} & LiDAR, visual camera & 3D Car & LiDAR point clouds, (processed by PointNet~\cite{qi2017pointnet}); RGB image (processed by a 2D CNN) & R-CNN & A 3D object detector for RGB image & After RP & Using RP from RGB image detector to search LiDAR point clouds & Late & KITTI \\ \midrule

Schneider \textit{et al.}, 2017\cite{schneider2017multimodal} & Visual camera & Multiple 2D objects & RGB image (processed by GoogLeNet), depth image from stereo camera (processed by NiN net) & SSD & SSD predictions & Before RP & Feature concatenation & Early, \textbf{Middle}, Late & Cityscape \\ \midrule

Takumi \textit{et al.}, 2017\cite{takumi2017multispectral} & Visual camera, thermal camera & Multiple 2D objects & RGB image, NIR, FIR, FIR image. Each processed by YOLO & YOLO & YOLO predictions for each spectral image & After RP & Ensemble: ensemble final predictions for each YOLO detector & Late & self-recorded data\\ \midrule

Chen \textit{et al.}, 2017\cite{chen2017multi} & LiDAR, visual camera & 3D Car & LiDAR BEV and spherical maps, RGB image. Each processed by a base network built on VGG16 & Faster R-CNN & A RPN from LiDAR BEV map & After RP & average mean, deep fusion & Early, \textbf{Middle}, Late & KITTI \\  \midrule

Asvadi \textit{et al.}, 2017\cite{asvadi2017multimodal} & LiDAR, visual camera & 2D Car & LiDAR front-view dense-depth (DM) and reflectance maps (RM), RGB image. Each processed through a YOLO net & YOLO & YOLO outputs for LiDAR DM and RM maps, and RGB image & After RP & Ensemble: feed engineered features from ensembled bounding boxes to a network to predict scores for NMS & Late & KITTI \\  \midrule

Oh \textit{et al.}, 2017\cite{oh2017object} & LiDAR, visual camera & 2D Car, Pedestrian, Cyclist & LiDAR front-view dense-depth map (for fusion: processed by VGG16), LiDAR voxel (for ROIs: segmentation and region growing), RGB image (for fusion: processed by VGG16; for ROIs: segmentation and grouping) & R-CNN & LiDAR voxel and RGB image separately & After RP & Association matrix using basic belief assignment & Late  & KITTI \\  \midrule

Wang \textit{et al.}, 2017\cite{wang2017fusing} & LiDAR, visual camera & 3D Car, Pedestrian  & LiDAR BEV map, RGB image. Each processed by a RetinaNet~\cite{lin2018focal} & One stage detector & Fused LiDAR and RGB image features extracted from CNN & Before RP & Sparse mean manipulation & Middle & KITTI \\  \midrule

Ku \textit{et al.}, 2017\cite{ku2017joint} & LiDAR, visual camera & 3D Car, Pedestrian, Cyclist  & LiDAR BEV map, RGB image. Each processed by VGG16 & Faster R-CNN & Fused LiDAR and RGB image features extracted from CNN & Before and after RP & Average mean & \textbf{Early}, Middle, Late & KITTI \\  \midrule

Xu \textit{et al.}, 2017\cite{xu2017pointfusion} & LiDAR, visual camera & 3D Car, Pedestrian, Cyclist, Indoor objects  & LiDAR points (processed by PointNet), RGB image (processed by ResNet) & R-CNN & Pre-trained RGB image detector & After RP & Feature concatenation for local and global features & Middle & KITTI, SUN-RGBD \\  \midrule

Qi \textit{et al.}, 2017\cite{qi2017frustum} & LiDAR, visual camera & 3D Car, Pedestrian, Cyclist, Indoor objects  & LiDAR points (processed by PointNet), RGB image (using a pre-trained detector) & R-CNN & Pre-trained RGB image detector & After RP & Feature concatenation & Middle, \textbf{Late} & KITTI, SUN-RGBD \\ \midrule

Du \textit{et al.}, 2017\cite{du2017car} & LiDAR, visual camera & 2D Car & LiDAR voxel (processed by RANSAC and model fitting), RGB image (processed by VGG16 and GoogLeNet) & Faster R-CNN & First clustered by LiDAR point clouds, then fine-tuned by a RPN of RGB image & Before RP & Ensemble: feed LiDAR RP to RGB image-based CNN for final prediction  & Late & KITTI \\ \midrule

Matti \textit{et al.}, 2017\cite{matti2017combining} & LiDAR, visual camera & 2D Pedestrian & LiDAR points (clustering with DBSCAN) and RGB image (processed by ResNet) & R-CNN & Clustered by LiDAR point clouds, then size and ratio corrected on RGB image.  & Before and at RP & Ensemble: feed LiDAR RP to RGB image-based CNN for final prediction & Late & KITTI \\ \midrule

Kim \textit{et al.}, 2016\cite{kim2016robust} & LiDAR, visual camera & 2D Pedestrian, Cyclist & LiDAR front-view depth image, RGB image. Each processed by Fast R-CNN network~\cite{girshick2015fast} & Fast R-CNN & Selective search for LiDAR and RGB image separately. & At RP & Ensemble: joint RP are fed to RGB image based CNN. & Late & KITTI \\ \midrule

Mees \textit{et al.}, 2016\cite{mees2016choosing} & RGB-D camera & 2D Pedestrian & RGB image, depth image from depth camera, optical flow. Each processed by GoogLeNet & Fast R-CNN & Dense multi-scale sliding window for RGB image & After RP & Mixture of Experts & Late & RGB-D People Unihall Dataset, InOutDoor RGB-D People Dataset. \\ \midrule

Wagner \textit{et al.}, 2016\cite{wagner2016multispectral} & visual camera, thermal camera & 2D Pedestrian & RGB image, thermal image. Each processed by CaffeeNet & R-CNN & ACF+T+THOG detector & After RP & Feature concatenation & Early, \textbf{Late} & KAIST Pedestrian Dataset \\ \midrule

Liu \textit{et al.}, 2016\cite{liu2016bmvc} & Visual camera, thermal camera & 2D Pedestrian & RGB image, thermal image. Each processed by NiN network & Faster R-CNN & RPN with fused (or separate) features & Before and after RP & Feature concatenation, average mean, Score fusion (Cascaded CNN) & Early, \textbf{Middle}, Late & KAIST Pedestrian Dataset \\ \midrule

Schlosser \textit{et al.}, 2016\cite{schlosser2016fusing} & LiDAR, visual camera & 2D Pedestrian  & LiDAR HHA image, RGB image. Each processed by a small ConvNet & R-CNN & Deformable Parts Model with RGB image & After RP & Feature concatenation & Early, Middle, Late & KITTI \\ 
\bottomrule
\end{tabularx}
\begin{tablenotes}
\item[a] For one-stage detector, we refer region proposals to be the detection outputs of a network.
\item[b] Some methods compare multiple fusion levels. We mark the fusion level with the best reported performance in bold.
\item[c] Besides object detection, this paper also proposes intention prediction and trajectory prediction up to 3s in the unified network (multi-task prediction).
\end{tablenotes}

\end{ThreePartTable}
\captionsetup{font=normalsize}
\scriptsize
\centering
\begin{threeparttable}
\begin{tabularx}{1.00\linewidth}{>{\raggedright\arraybackslash\hsize=0.3\hsize}X >{\raggedright\arraybackslash\hsize=0.4\hsize}X >{\raggedright\arraybackslash\hsize=0.4\hsize}X >{\raggedright\arraybackslash\hsize=0.8\hsize}X >{\raggedright\arraybackslash\hsize=0.7\hsize}X >{\raggedright\arraybackslash\hsize=0.4\hsize}X >{\raggedright\arraybackslash\hsize=0.3\hsize}X}
\caption{SUMMARY OF MULTI-MODAL SEMANTIC SEGMENTATION METHODS}\label{tab:object_segementation_summary}\\
\rowcolor{lightgray!75}
\textbf{Reference} & \textbf{Sensors} & \textbf{Semantics}& \textbf{Sensing Modality Representations} & \textbf{Fusion Operation and Method} & \textbf{Fusion Level \tnote{a}} & \textbf{Dataset(s) used} \\[4pt]

\endhead
Chen \textit{et al.}, 2019\cite{chen2019progressive} & LiDAR, visual camera & Road segmentation & RGB image, altitude difference image. Each processed by a CNN & Feature adaptation module, modified concatenation. & Middle & KITTI \\ \midrule

Valada \textit{et al.}, 2019\cite{valada2018self} & Visual camera, depth camera, thermal camera  & Multiple 2D objects & RGB image, thermal image, depth image. Each processed by FCN with ResNet backbone (Adapnet++ architecture) & Extension of Mixture of Experts & Middle & Six datasets, including Cityscape, Sun RGB-D, etc. \\ \midrule

Sun \textit{et al.}, 2019\cite{sun2019rtfnet} & Visual camera, thermal camera & Multiple 2D objects in campus environments & RGB image, thermal image. Each processed by a base network built on ResNet & Element-wise summation in the encoder networks & Middle & Datasets published by~\cite{ha2017mfnet} \\ \midrule

Caltagirone \textit{et al.}, 2019\cite{caltagirone2019lidar} & LiDAR, visual camera & Road segmentation & LiDAR front-view depth image, RGB image. Each input processed by a FCN & Feature concatenation (For early and late fusion), weighted addition similar to gating network (for middle-level cross fusion) & Early, \textbf{Middle}, Late & KITTI \\ \midrule

Erkent \textit{et al.}, 2018\cite{erkent2018semantic} & LiDAR, visual camera & Multiple 2D objects & LiDAR BEV occupancy grids (processed based on Bayesian filtering and tracking), RGB image (processed by a FCN with VGG16 backbone) & Feature concatenation & Middle & KITTI, self-recorded \\ \midrule

Lv \textit{et al.}, 2018\cite{lv2018novel} & LiDAR, visual camera & Road segmentation & LiDAR BEV maps, RGB image. Each input processed by a FCN with dilated convolution operator. RGB image features are also projected onto LiDAR BEV plane before fusion & Feature concatenation  & Middle & KITTI \\ \midrule

Wulff \textit{et al.}, 2018\cite{wulff2018early} & LiDAR, visual camera & Road segmentation. Alternatives: freespace, ego-lane detection & LiDAR BEV maps, RGB image projected onto BEV plane. Inputs processed by a FCN with UNet & Feature concatenation & Early & KITTI \\ \midrule

Kim \textit{et al.}, 2018\cite{kim2018season} & LiDAR, visual camera & 2D Off-road terrains  & LiDAR voxel (processed by 3D convolution), RGB image (processed by ENet) & Addition & Early, \textbf{Middle}, Late & self-recorded data \\ \midrule

Guan \textit{et al.}, 2018\cite{guan2018fusion}\tnote{b} & Visual camera, thermal camera & 2D Pedestrian & RGB image, thermal image. Each processed by a base network built on VGG16 & Feature concatenation, Mixture of Experts & Early, Middle, Late & KAIST Pedestrian Dataset \\ \midrule

Yang \textit{et al.}, 2018\cite{yang2018fusion} & LiDAR, visual camera & Road segmentation & LiDAR points (processed by PointNet++), RGB image (processed by FCN with VGG16 backbone) & Optimizing Conditional Random Field (CRF) & Late & KITTI \\ \midrule

Gu \textit{et al.}, 2018\cite{gu20183d} & LiDAR, visual camera & Road segmentation & LiDAR front-view depth and height maps (processed by a inverse-depth histogram based line scanning strategy), RGB image (processed by a FCN). & Optimizing Conditional Random Field & Late & KITTI \\ \midrule

Cai \textit{et al.}, 2018\cite{cai2018robust} & Satellite map with route information, visual camera & Road segmentation & Route map image, RGB image. Images are fused and processed by a FCN & Overlaying the line and curve segments in the route map onto the RGB image to generate the Map Fusion Image (MFI) & Early & self-recorded data \\ \midrule

Ha \textit{et al.}, 2017\cite{ha2017mfnet} & Visual camera, thermal camera & Multiple 2D objects in campus environments & RGB image, thermal image. Each processed by a FCN and mini-inception block & Feature concatenation, addition (``short-cut fusion'') & Middle & self-recorded data \\ \midrule

Valada \textit{et al.}, 2017\cite{valada2017adapnet} & Visual camera, thermal camera & Multiple 2D objects & RGB image, thermal image, depth image. Each processed by FCN with ResNet backbone & Mixture of Experts & Late & Cityscape, Freiburg Multispectral Dataset, Synthia \\ \midrule

Schneider \textit{et al.}, 2017\cite{schneider2017multimodal} & Visual camera & Multiple 2D Objects  & RGB image, depth image & Feature concatenation & Early, \textbf{Middle}, Late & Cityscape \\ \midrule

Schneider \textit{et al.}, 2017\cite{schneider2017multimodal}\tnote{2} & Visual camera & Multiple 2D Objects  & RGB image (processed by GoogLeNet), depth image from stereo camera (processed by NiN net) & Feature concatenation & Early, \textbf{Middle}, Late & Cityscape \\ \midrule

Valada \textit{et al.}, 2016\cite{valada2016deep} & Visual camera, thermal camera & Multiple 2D objects in forested environments & RGB image, thermal image, depth image. Each processed by the UpNet (built on VGG16 and up-convolution) & Feature concatenation, addition & Early, \textbf{Late} & self-recorded data \\ 
\bottomrule
\end{tabularx}

\begin{tablenotes}
\item[a] Some methods compare multiple fusion levels. We mark the fusion level with the best reported performance in bold.
\item[b] They also test the methods for object detection problem with different network architectures (see Table~\ref{tab:object_detection_summary}). 
\end{tablenotes}
\end{threeparttable}

\captionsetup{font=normalsize}
\scriptsize
\centering
\begin{ThreePartTable}
	\begin{tabularx}{1.0\linewidth}{
	>{\raggedright\arraybackslash\hsize=0.8\hsize}X
	>{\raggedright\arraybackslash\hsize=0.5\hsize}X
	>{\raggedright\arraybackslash\hsize=0.5\hsize}X 
	>{\raggedright\arraybackslash\hsize=0.5\hsize}X 
	>{\raggedright\arraybackslash\hsize=0.5\hsize}X
	>{\raggedright\arraybackslash\hsize=0.5\hsize}X
	>{\raggedright\arraybackslash\hsize=0.5\hsize}X
	>{\raggedright\arraybackslash\hsize=0.5\hsize}X
	>{\raggedright\arraybackslash\hsize=0.5\hsize}X
	>{\raggedright\arraybackslash\hsize=0.5\hsize}X
	>{\raggedright\arraybackslash\hsize=0.2\hsize}X  >{\hsize=0.0\hsize}X}
		\caption{PERFORMANCE AND RUNTIME FOR 3D OBJECT DETECTION ON KITTI TEST SET}\label{tab:perf_3d}\\
		\rowcolor{lightgray!75}
		\multicolumn{1}{c}{\textbf{Reference}} & \multicolumn{3}{c}{\textbf{Car}}  & 
		\multicolumn{3}{c}{\textbf{Pedestrian}}  &
		\multicolumn{3}{c}{\textbf{Cyclist}}  &
		\multicolumn{1}{c}{\textbf{Runtime}} & \multicolumn{1}{c}{\textbf{Environment}} \\
		\rowcolor{lightgray!75}
		 & \textbf{Moderate} & \textbf{Easy} & \textbf{Hard} & \textbf{Moderate} & \textbf{Easy} & \textbf{Hard} & \textbf{Moderate} & \textbf{Easy} & \textbf{Hard} & & \\[6pt]
		\endhead
		\endfoot
\toprule

Liang \textit{et al.}, 2019\cite{liang2019multi} &  \textbf{76.75 \%} &  \textbf{86.81 \%} &  \textbf{68.41 \%} &  \textbf{45.61 \%} &  \textbf{52.37 \%} &  \textbf{41.49 \%} &  \textbf{64.68 \%} &  \textbf{79.58 \%} &  \textbf{57.03 \%} & 0.08 s & GPU @ 2.5 Ghz (Python) \\
Wang \textit{et al.}, 2019\cite{wang2019frustum} &  76.51 \% &  85.88 \% &  68.08 \% & - & - & - & - & - & - & 0.47 s & GPU @ 2.5 Ghz (Python + C/C++) \\
Sindagi \textit{et al.}, 2019\cite{sindagi2019mvx} &  72.7 \% &  83.2 \% &  65.12 \% & - & - & - & - & - & - & - & -  \\
Shin \textit{et al.}, 2018\cite{shin2018roarnet} &  73.04 \% &  83.71 \% &  59.16 \% & - & - & - & - & - & - & - & GPU Titan X (not Pascal) \\
Du \textit{et al.}, 2018\cite{du2018general} &  73.80 \% &  84.33 \% &  64.83 \% & - & - & - & - & - & - & 0.5 s & GPU @ 2.5 Ghz (Matlab + C/C++) \\
Liang \textit{et al.}, 2018\cite{liang2018deep} &  66.22 \% &  82.54 \% &  64.04 \% & - & - & - & - & - & - & \textbf{0.06 s} & GPU @ 2.5 Ghz (Python) \\
Ku \textit{et al.}, 2017\cite{ku2017joint} &  71.88 \% &  81.94 \% &  66.38 \% &  42.81 \% &  50.80 \% &  40.88 \% &  52.18 \% &  64.00 \% &  46.61 \% & 0.1 s & GPU Titan X (Pascal) \\
Qi \textit{et al.}, 2017\cite{qi2017frustum} &  70.39 \% &  81.20 \% &  62.19 \% &  44.89 \% &  51.21 \% &  40.23 \% &  56.77 \% &  71.96 \% &  50.39 \% & 0.17 s & GPU @ 3.0 Ghz (Python) \\
Chen \textit{et al.}, 2017\cite{chen2017multi} &  62.35 \% &  71.09 \% &  55.12 \% & - & - & - & - & - & - & 0.36 s & GPU @ 2.5 Ghz (Python + C/C++) \\
\bottomrule
\end{tabularx}
\end{ThreePartTable}

\begin{ThreePartTable}
	\begin{tabularx}{1.0\linewidth}{
	>{\raggedright\arraybackslash\hsize=0.8\hsize}X
	>{\raggedright\arraybackslash\hsize=0.5\hsize}X
	>{\raggedright\arraybackslash\hsize=0.5\hsize}X 
	>{\raggedright\arraybackslash\hsize=0.5\hsize}X 
	>{\raggedright\arraybackslash\hsize=0.5\hsize}X
	>{\raggedright\arraybackslash\hsize=0.5\hsize}X
	>{\raggedright\arraybackslash\hsize=0.5\hsize}X
	>{\raggedright\arraybackslash\hsize=0.2\hsize}X  >{\hsize=0.0\hsize}X}
		\caption{PERFORMANCE AND RUNTIME FOR ROAD SEGMENTATION (URBAN) ON KITTI TEST SET}\label{tab:perf_semseg}\\
		\rowcolor{lightgray!75}
		\textbf{Method} & \textbf{MaxF} & \textbf{AP} & \textbf{PRE} & \textbf{REC} & \textbf{FPR} & \textbf{FNR} & \textbf{Runtime} & \textbf{Environment} \\[6pt]
		\endhead
		\endfoot
\toprule
Chen \textit{et al.}, 2019\cite{chen2019progressive} & \textbf{97.03} \% & \textbf{94.03} \% & \textbf{97.19} \% & \textbf{96.88} \% & \textbf{1.54} \% & \textbf{3.12} \% & 0.16 s & GPU \\
Caltagirone \textit{et al.}, 2019\cite{caltagirone2019lidar} & 96.03 \% & 93.93 \% & 96.23 \% & 95.83 \% & 2.07 \% & 4.17 \% & 0.15 s & GPU \\
Gu \textit{et al.}, 2018\cite{gu20183d} & 95.22 \% & 89.31 \% & 94.69 \% & 95.76 \% & 2.96 \% & 4.24 \% & \textbf{0.07 s} & CPU \\
Lv \textit{et al.}, 2018\cite{lv2018novel} & 94.48 \% & 93.65 \% & 94.28 \% & 94.69 \% & 3.17 \% & 5.31 \% & - & GPU Titan X \\
Yang \textit{et al.}, 2018\cite{yang2018fusion} & 91.40 \% & 84.22 \% & 89.09 \% & 93.84 \% & 6.33 \% & 6.16 \% & - & GPU \\
\bottomrule
\end{tabularx}
\end{ThreePartTable}
\clearpage
\end{landscape}
\end{document}